\documentclass[conference]{IEEEtran}
\usepackage{times}

\usepackage[numbers]{natbib}
\usepackage{multicol}
\usepackage{amsmath}
\usepackage[bookmarks=true]{hyperref}
\usepackage{siunitx}
\usepackage{amssymb}
\usepackage{booktabs}
\usepackage{float}
\let\labelindent\rela
\usepackage{enumitem}
\usepackage{pgfplots}
\usepackage{pgfplotstable}
\pgfplotsset{compat=1.18}
\usepackage{xcolor}
\definecolor{tabblue}{rgb}{0.121568627, 0.466666667, 0.705882353}
\definecolor{taborange}{rgb}{1.0, 0.498039216, 0.0}
\usepackage{multirow}

\begin{document}

\title{Accelerating Model-Based Reinforcement Learning with State-Space World Models}

\author{\authorblockN{Maria Krinner\authorrefmark{1}, Elie Aljalbout\authorrefmark{1}, Angel Romero, Davide Scaramuzza} 
\authorblockA{
Robotics and Perception Group, University of Zurich, Switzerland\\
\authorrefmark{1}These authors contributed equally\vspace{15pt}
}
}

\maketitle

\begin{abstract}
Reinforcement learning~(RL) is a powerful approach for robot learning.
However, model-free RL~(MFRL) requires a large number of environment interactions to learn successful control policies.
This is due to the noisy RL training updates and the complexity of robotic systems, which typically involve highly non-linear dynamics and noisy sensor signals.
In contrast, model-based RL (MBRL) not only trains a policy but simultaneously learns a world model that captures the environment's dynamics and rewards.
The world model can either be used for planning, for data collection, or to provide first-order policy gradients for training.
Leveraging a world model significantly improves sample efficiency compared to model-free RL. However, training a world model alongside the policy increases the computational complexity, leading to longer training times that are often intractable for complex real-world scenarios.
In this work, we propose a new method for accelerating model-based RL using state-space world models.
Our approach leverages state-space models (SSMs) to parallelize the training of the dynamics model, which is typically the main computational bottleneck.
Additionally, we propose an architecture that provides privileged information to the world model during training, which is particularly relevant for partially observable environments.
We evaluate our method in several real-world agile quadrotor flight tasks, involving complex dynamics, for both fully and partially observable environments. 
We demonstrate a significant speedup, reducing the world model training time by up to 10 times, and the overall MBRL training time by up to 4 times.
This benefit comes without compromising performance, as our method achieves similar sample efficiency and task rewards to state-of-the-art MBRL methods.

\end{abstract}

\IEEEpeerreviewmaketitle

\newcommand{\Dist}[4]{#1_{#2}\left(#3 \,\middle|\, #4\right)}
\newcommand{\KL}[2]{D_{\text{KL}}\left[#1 \,\middle\|\, #2\right]}
\newcommand{\sg}{\ensuremath{\operatorname{sg}}}

\section{Introduction}
\label{sec:intro}

    Robot learning has proven to be a very successful paradigm for acquiring complex robotic skills. 
    Recent work has demonstrated the applicability of reinforcement learning~(RL) approaches to manipulation~\cite{andrychowicz2020learning,handa2023dextreme,aljalbout2024role,o2024open}, locomotion~\cite{tan2018sim, hwangbo2019learning}, and aerial robotics~\cite{kaufmann23champion,song2023reaching}.
    For such methods to succeed, they either require encoding the task into a reward function or access to a dataset of expert demonstrations.
    Learning control policies reduces the barrier to skill acquisition in robotics, requiring less human interventions and efforts to equip a robot with a new skill.
    In addition, RL methods impose minimal constraints in the design of reward functions.
    As a result, they allow for greater flexibility in the design of robotic systems, which can lead to outperforming classical control approaches~\cite{song2023reaching, lee2020learning}.
    
    Despite their success, one major challenge in robot learning is the need for large amounts of physical interaction data, which can be very expensive and challenging to obtain. 
    In addition, RL suffers from high-variance gradients, often leading to unstable training and raising the need for even more data samples.
    This problem is further exacerbated when learning vision-based control policies due to the high-dimensionality of image-based observations.

    \begin{figure}[t!]
        \centering
        \includegraphics[width=\linewidth]{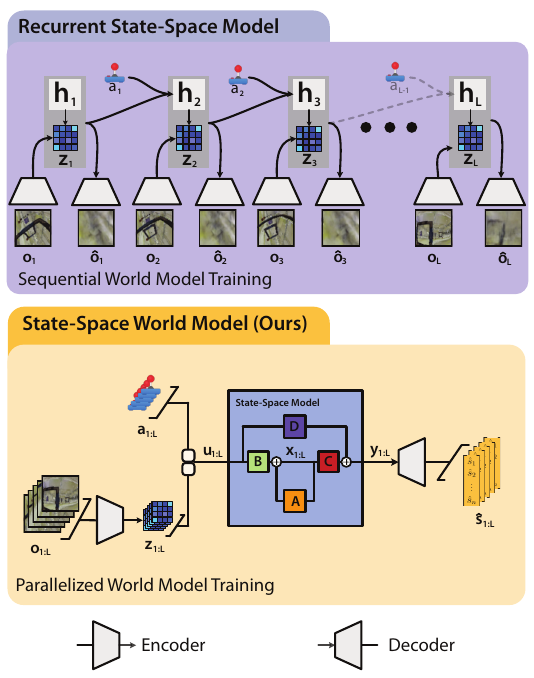}
        \caption{State-of-the-art model-based RL~(MBRL) methods typically employ recurrent state-space models (RSSMs) as the world model backbone, which are slow in training due to the sequential nature of RNNs.
        We leverage state-space models (SSMs) to parallelize the sequence dimension of the world model, thereby reducing the computational complexity of training the world model. 
        Moreover, we propose reconstructing privileged observations of lower dimensionality $s_t$, rather than high-dimensional image observations $o_t$.}
        \label{fig:fig1}
    \end{figure}
    
    To alleviate this problem, multiple solutions have been proposed to leverage prior knowledge in RL.
    These solutions range from pretraining the policy or parts of it using either imitation learning~\cite{xing2024bootstrapping} or self-supervised learning objectives~\cite{yarats2021improving, aljalbout2021learning,lee2020making}, to embedding inductive biases into RL policy architectures and training pipelines to reduce the policy complexity~\cite{lutter2021differentiable,funk2022learn2assemble}.
    Model-based RL has emerged as a promising alternative to improve sample efficiency in comparison to model-free RL~\cite{deisenroth2011learning, heess2015learning,Hafner2020Dream, aljalbout2024limt}.
    
    However, model-based RL methods are typically slower in training than their model-free counterparts because they additionally train the world model~(WM), which can be slow due to the sequential nature of the dynamics.
    This aspect limits the applicability of MBRL, particularly for cases where fast training is desirable.
    
    In this work, we propose a method for accelerating model-based reinforcement learning~(MBRL).
    We leverage state-space models~(SSMs) to parallelize the sequence dimension of the world model, thereby reducing the computational complexity of MBRL.
   We build on the family of Dreamer-based MBRL methods~\cite{Hafner2020Dream, hafner2023mastering}.
    In our approach, we replace the recurrent state-space model~(RSSM) with a modern parallelizable SSM as the dynamics model.
    
    To evaluate our approach, we perform experiments in a drone racing environment, involving complex dynamics, for both fully and partially observable environments. We then compare our method to state-of-the-art model-free and model-based RL methods.
    Our method achieves a significant speedup, reducing the overall training time by up to 4 times.
    This benefit comes without compromising performance, as our method attains similar sample efficiency and task rewards as state-of-the-art methods.

\subsection*{Contributions}
Our contributions are summarized as follows:
\begin{itemize}
    \item We present a method for leveraging state-space models to accelerate the sequence model training in MBRL world models.
    \item Our method achieves comparable sample efficiency to state-of-the-art MBRL methods while reducing world model training time by up to 10 times, and the overall MBRL training time by up to 4 times.
    \item We present an approach to facilitate the sim-to-real transfer of our vision-based policies that leverage the privileged state information while training the vision-based world model.
    \item We demonstrate our approach in the real-world robotic task of agile quadrotor flight, which involves nonlinear complex dynamics.%
\end{itemize}

\section{Related Work}
\label{sec:relatedwork}

\subsection{Model-Based RL}
While model-free RL methods, such as proximal policy optimization~(PPO), have been prevalent in robot learning~\cite{hwangbo2019learning, andrychowicz2020learning,alles2022learning,song2023reaching, handa2023dextreme,aljalbout2024role}, model-based RL is gaining popularity due to its sample efficiency.
The main difference between the two paradigms is that MBRL learns a model of the environment's dynamics, in addition to the policy~\cite{sutton2018reinforcement}.
This model, known as world model, can either be used for i) planning and control~\cite{nagabandi2018neural, hansen2022temporal}, ii) sampling environment interactions for policy training~\cite{janner2019trust,yu2020mopo,kidambi2020morel}, or iii) obtaining first-order policy gradients through imagination~\cite{heess2015learning,Hafner2020Dream,hafner2023mastering}.
World models play a crucial role in MBRL, with recurrent neural networks (RNNs) being the most widely adopted architecture. 
One popular RNN-based world model architecture is the Recurrent State-Space Model (RSSM), introduced in the Dreamer framework~\cite{hafner2019learning}.
However, RSSMs struggle to scale efficiently to long sequences, as their computational complexity increases significantly with the sequence length.

Recent works have explored alternative architectures, such as Transformers and SSMs. 
Transformers, in particular, have been widely adopted as world model backbones in MBRL, showing advantages both for sample efficiency and computational complexity~\cite{robine2023twm, zhang2023storm, iris2023, chen2022transdreamer}.
Similarly, in~\cite{deng2024facing} an S4-based world model is introduced.
However, this work does not explore the usage of such world models for MBRL and focuses on mastering predictions in long-range memory tasks. 

One of the earliest applications of MBRL to robotics was shown on a low-cost manipulator and using a Gaussian process-driven policy representation~\cite{deisenroth2011learning}.
Later work adopted a combination of deep dynamics models with model predictive control to scale this concept to vision-based robotics tasks~\cite{Wahlstrom2015c}.
\citet{nagabandi2018neural} proposed using MBRL to gather expert data for training an initial policy using model-free methods.
By doing so, they managed to alleviate the inferior task performance of MBRL methods.
\citet{chua2018deep} proposed using ensembles of dynamics model to enable uncertainty propagation during trajectory sampling.
These approaches have also been extended to jointly learn a value function together with the dynamics sequence model~\cite{hansen2022temporal,hansen2024tdmpc}.
Most of these methods perform receding horizon control with a sampling scheme.

Another line of work leverages first-order policy gradients that are backpropagated through a learned latent dynamics model~\cite{heess2015learning,Hafner2020Dream,hafner2023mastering}, which has been successfully applied to multiple robotic tasks in manipulation, locomotion, and drone flight~\cite{wu2023daydreamer, becker2020learning,brunnbauer2022latent,richard2022learning,aljalbout2024limt,bi2024sample,yamada2024twist}.

\subsection{Vision-Based Robot Learning}
While state-based RL can be challenging, learning vision-based policies presents additional difficulties. First, the higher dimensionality of the observation space makes policy search more complex, as the model must process significantly larger inputs. Second, tasks relying on visual inputs often suffer from partial observability, where the raw input does not provide a complete representation of the state. Therefore, such policies typically require additional supervision signals to effectively learn informative latent representations from raw pixel data. As a result, vision-based RL often requires a large amount of data to achieve good performance, as individual samples tend to be less informative compared to state-based environments~\cite{lillicrap2015continuous, kalashnikov2018scalable}.

Several strategies have been proposed to improve the sample-efficiency of vision-based RL. One prominent approach is to embed auxiliary losses to provide additional signals for training the perception components of the policy.
Auxiliary losses are typically inspired by the self-supervised learning literature and include methods like auto-encoding and contrastive learning~\cite{yarats2021improving,laskin2020curl}.
These methods were mostly successful in simulation~\cite{yarats2021improving,laskin2020curl,aljalbout2021learning}, but have also been deployed in real-world scenarios~\cite{chen2021robust,xing2024contrastive,silwal2024we,van2016stable}.
Motivated by the success of imitation learning for vision-based policy learning~\cite{fu2024mobile,o2024open}, multiple efforts have been made to bootstrap RL with imitation learning~\cite{DBLP:conf/rss/Zhu0MRECTKHFH18,huflare,xing2024bootstrapping}.
Model-based RL offers a more generic approach that bypasses intermediate abstractions and directly learns a control policy from visual inputs in an end-to-end fashion. Recent works have shown impressive performance across a variety of manipulation, locomotion, and agile flight tasks in real-world scenarios~\cite{wu2023daydreamer, romero2024dream}.

\section{Preliminaries}

\subsection{Reinforcement Learning}
Given a reward function, RL  automates policy learning or search, by  maximizing the cumulative reward in a given environment~\cite{sutton2018reinforcement}.
Tasks are usually formulated as Markov Decision Processes~(MDP)~\cite{bellman1957markovian}. A finite-horizon, discounted MDP is defined by the tuple $\mathcal{M} = (\mathcal{S}, \mathcal{A}, \mathcal{P}, r, \rho_0, \gamma, T )$, where $\mathcal{S}$ and $\mathcal{A}$ are the state and action spaces respectively, $\mathcal{P}: \mathcal{S} \times \mathcal{A} \to \mathcal{S}$ is the transition dynamics, and $r: \mathcal{S} \times \mathcal{A} \to \mathbb{R}$ the reward.
Additionally, the MDP definition includes an initial state distribution $\rho_0$, a discount factor $\gamma \in [0,1]$, and a horizon length $T$. 
The optimal policy $\pi : \mathcal{S} \to P(\mathcal{A})$ maximizes the expected discounted reward as an objective function,
\begin{equation}
\label{eq:RL}
J(\pi) =  E_{\pi} \left[ \sum_{t=0}^{T-1} \gamma^t r(s_t, a_t) \right].
\end{equation}
In MBRL, in addition to learning the policy, we learn a model of the environment dynamics, usually referred to as world model of the form $\text{WM}: \mathcal{O} \times \mathcal{A} \to \mathcal{O} \times \mathbb{R}$.
The world model approximates the environment's transition dynamics $p(s_{t+1}\mid{s_t, a_t})$ and rewards $r(s_t, a_t)$ providing a simulated environment in which the agent can perform planning or optimization without requiring actual environment interactions.
This approach is often referred to as \textit{learning in imagination}.
Training typically follows an iterative process composed of three steps.
The first step is to train the world model using the collected data.
In the second step, we optimize the policy within the WM-based simulated environment.
In the third step, we collect new data by interacting with the actual environment using our latest policy.

\subsection{State-Space Models}
State space models (SSMs) provide an efficient framework for sequence modeling tasks and have shown great success in capturing long-term temporal and spatial dependencies~\cite{gu2022efficiently,smith2023simplified,patro2024mamba}. At time $t$, a first-order linear system maps the input $u(t) \in \mathbb{R}^H$ to the output $y(t) \in \mathbb{R}^P$ via the following system dynamics,
\begin{equation}
\begin{aligned}
\frac{dx(t)}{dt} = A x(t) + B u(t) \\
y(t) = C x(t) + D u(t),
\end{aligned}
\end{equation}
where $A, B, C, D$ are learnable parameters which capture the specific dependencies in the data, and $x(t)$ is the hidden state. To model sequences with a fixed time step $\Delta$, the system is discretized as follows,
\begin{equation}
\begin{aligned}
x_t = \bar{A} x_{t-1} + \bar{B} u_t \\
y_t = \bar{C} x_t + \bar{D} u_t,
\end{aligned}
\end{equation}
where $\bar{A}, \bar{B}, \bar{C}, \bar{D}$ are the discretized matrices, computed as functions of the continuous-time matrices $A, B, C, D$.

Given a sequence of inputs $e_{1:L}$, the parallel scan operation efficiently applies an associative binary operator, $\bullet$, to its inputs. Specifically, for a sequence of length $L$, the parallel scan computes:
\begin{equation}
\label{eq:scan}
[e_1, e_1 \bullet e_2, e_1 \bullet e_2 \bullet e_3, \dots, e_1 \bullet e_2 \bullet \dots \bullet e_L].
\end{equation}
Equation \ref{eq:scan} can be computed in $\mathcal{O}(\log(L))$ using $L$ parallel processors. S5~\cite{smith2023simplified} is an SSM that leverages the parallel scan to compute the sequence of hidden states $x_{1:L}$.
In the context of S5, the binary operator is defined as
\begin{equation}
\label{eq:associative}
a_i \bullet a_j = (a_{j,a} \odot a_{i,a}, a_{j,a} \otimes a_{i,b} + a_{j,b}),
\end{equation}
where $\odot$ represents matrix-matrix multiplication and $\otimes$ represents matrix-vector multiplication.
The operator is applied to the initial elements $e_k$, which are given by the pairs of matrices
\begin{equation}
\label{eq:sequence_scan}
e_k = (e_{k,a}, e_{k,b}) := (\bar{A}, \bar{B} u_k).
\end{equation}
Using this definition, the sequence of hidden states $x_{1:L}$ is computed iteratively as 
\begin{equation}
\label{eq:full_scan}
\begin{aligned}
e_1 &= (\bar{A}, \bar{B} u_1) = (\bar{A}, x_1) \\
e_1 \bullet e_2 &= (\bar{A}^2, \bar{A} x_1 + \bar{B} u_2) = (\bar{A}^2, x_2) \\
e_1 \bullet e_2 \bullet e_3 &= (\bar{A}^3, \bar{A} x_2 + \bar{B} u_3) = (\bar{A}^3, x_3).
\end{aligned}
\end{equation}
In general, matrix-matrix multiplications with a matrix $\bar{A}$ would incur a time complexity of $\mathcal{O}(P^3)$. However, by diagonalizing the system’s matrices, the complexity is reduced to $\mathcal{O}(P L)$, where $P$ is the size of the matrix and $L$ is the length of the sequence. As a result, the parallel scan operation is computed in $\mathcal{O}(\log(P L))$, making it scalable to long sequences.

\section{Methodology}
\label{sec:method}
In this work, we propose a model-based RL approach that leverages state-space world models. 
At each training step, we sample data from the replay buffer to train the world model as explained in section~\ref{sec:s5wm} and the actor-critic as explained in section~\ref{sec:ac}.

\subsection{State-Space World Model}
\label{sec:s5wm}
The proposed world model is based on DreamerV3~\cite{hafner2023mastering}.
However, we replace the recurrent state-space model~(RSSM)~\cite{hafner2019learning} with an SSM as the sequence model approximating the transition dynamics.
Our model includes an encoder for latent variable inference, which maps the observations $o_t$, to posterior stochastic representations $z_t$.
For a prediction horizon $H$, the sequence model predicts the deterministic representations $y_{t+1\,:\,t+H}$ given the corresponding actions $a_{t\,:\,t+H-1}$, and previous posterior $z_{t\,:\,t+H-1}$. We then infer the prior stochastic representation $\hat{z}_t$ from $y_t$.
In addition, the model includes reward and episode continuation predictors as well as an observation decoder,
\begin{equation}
\begin{alignedat}{4}
&\text{Encoder:} && \quad && z_t    && \sim \;  \Dist{q}{\phi}{\cdot}{o_t} \\
&\text{Sequence model:} && \quad x_t, \; && y_t && = \; \text{SSM}(z_{t-1}, a_{t-1}, c_{t-1}, x_{t-1})  \\
&\text{Dynamics:} && \quad && \hat{z}_t  && \sim \; \Dist{p}{\phi}{\cdot}{y_t} \\
&\text{Decoder:} && \quad && \hat{o}_t   && \sim \; \Dist{p}{\phi}{\cdot}{y_t, z_t} \\
&\text{Reward prediction:} && \quad && \hat{r}_t && \sim \; \Dist{p}{\phi}{\cdot}{y_t, z_t} \\
&\text{Continue prediction:} && \quad && \hat{c}_t && \sim\; \Dist{p}{\phi}{\cdot}{y_t, z_t}.
\end{alignedat}
\label{eq:wm}
\end{equation}
We fuse $z_t$ and $a_t$ using a multi-layer perceptron~(MLP) which outputs a single vector $u_t$. 
We then feed $u_t$ as input to the SSM Block (see Appendix~\ref{sec:s5block}).

Unlike RSSMs, which process the sequence one step at a time, SSMs parallelize over the sequence dimension, making them particularly efficient for training on long sequences.
Among the different SSM architectures, we specifically choose S5~\cite{smith2023simplified}, which employs the parallel scan operation.
One key advantage of S5 is its ability to reset the hidden state, which is not possible in other SSMs, such as S4~\cite{gu2022efficiently}.
This is crucial for preventing the model from carrying over irrelevant information from previous episodes, which is particularly relevant in scenarios where sequence boundaries or discontinuities must be handled.
We leverage the associative operator introduced in~\cite{lu2024structured}, which enables efficient resetting of the hidden state.
We extend the initial elements $e_k$ from equation~\ref{eq:sequence_scan} to include the continuity predictions $c_k\in\{0,1\}$,
\begin{equation}
\label{eq:sequence_scan_reset}
e_k = (e_{k,a}, e_{k,b}, e_{k,c}) := (\bar{A}, \bar{B} u_k, 1-c_k),
\end{equation}
where $e_{k,c}$ represents whether the episode is done.
We then adapt the binary operator from~\ref{eq:associative} to incorporate the continuity predictors,
\begin{equation}
\label{eq:associative_reset}
a_i \bullet a_j = 
\begin{cases}
(a_{j,a} \odot a_{i,a}, a_{j,a} \otimes a_{i,b} + a_{j,b}, a_{i,c}) & \text{if } a_{j,c} = 0, \\
(a_{j,a}, a_{j,b}, a_{j,c}) & \text{if } a_{j,c} = 1.
\end{cases}
\end{equation}
Using these definitions, we compute the sequence of hidden states $x_{1:L}$ as in equation~\ref{eq:full_scan}.
Since our world model uses an S5 backbone for sequence modeling, we refer to our method as S5WM.

For optimizing the world model, we use the same loss function as in~\cite{hafner2023mastering}, which is derived based on the evidence lower bound of the marginal data likelihood.
The loss combines a prediction loss $\mathcal{L}_{pred}$, a dynamic loss $\mathcal{L}_{dyn}$, and a representation loss $\mathcal{L}_{rep}$,
\begin{small}
\begin{equation}
    \mathcal{L}(\phi) \doteq  \mathbb{E}_{q_\phi}\biggl[ \sum_{t=1}^T \beta_{pred} \mathcal{L}_{pred}(\phi) + \beta_{dyn} \mathcal{L}_{dyn}(\phi) \\
      +\beta_{rep} \mathcal{L}_{rep}(\phi) \biggr],
\end{equation}
\end{small}
where $\beta_{pred}$, $\beta_{dyn}$, and $\beta_{rep}$ are hyperparameters that modulate the effects of the different loss components,
\begin{equation}
\begin{aligned}
    \mathcal{L_{\mathrm{pred}}}(\phi) & \doteq
-\ln \Dist{p}{\phi}{o_t}{y_t, z_t}
-\ln \Dist{p}{\phi}{r_t}{y_t, z_t}
\\
& \hphantom{=} \,\,
-\ln \Dist{p}{\phi}{c_t}{y_t, z_t} \\
\mathcal{L_{\mathrm{dyn}}}(\phi) & \doteq
\max\bigl(1, \KL{\sg(\Dist{q}{\phi}{z_t}{o_t})}{\Dist{p}{\phi}{z_t}{x_t}\hphantom{)}}\bigr) \\
\mathcal{L_{\mathrm{rep}}}(\phi) & \doteq
\max\bigl(1, \KL{\Dist{q}{\phi}{z_t}{o_t})}{ \sg(\Dist{p}{\phi}{z_t}{x_t})}\bigr),
\end{aligned}
\end{equation}
where $D_{\text{KL}}$ refers to the Kullback-Leibler divergence, and $\sg$ to the stop gradient operation.

Each component of $\mathcal{L}(\phi)$ contributes to training a different component of the state-space world model.
The representation loss ensures that the encoder learns an informative posterior stochastic latent $\Dist{q}{\phi}{z_t}{o_t}$.
The dynamic loss helps the world model learn the transition dynamics by encouraging the prior stochastic latent $\Dist{p}{\phi}{z_t}{x_t}$, to align with the posterior.
The prediction loss trains the decoder, as well as the reward and continue prediction heads, guiding the model to accurately predict next states.
\begin{figure}%
        \centering
        \includegraphics[width=\linewidth]{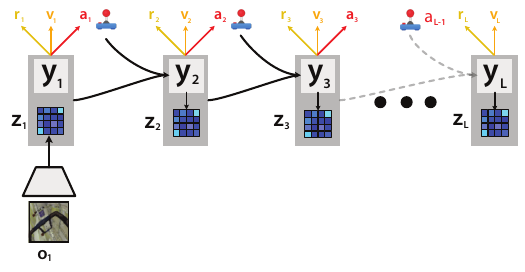}
        \caption{To train the actor and critic, we leverage imaginations in the latent state and using our state-space world model.
        We obtain these imaginations by encoding the initial observation and rolling out the sequence model in the latent space.
        Despite the possibility of training the world model in a parallel fashion, the imagination step (needed to train the actor-critic) cannot be parallelized due to the dependence on the policy to generate actions needed for rolling out the trajectories.}
        \label{fig:imaginationprocess}
\end{figure}
\subsection{Actor-Critic Training}
\label{sec:ac}

In practice, the world model serves as an approximate differentiable simulator providing first-order gradients for policy updates~\cite{xu2022accelerated}. This helps reduce variance estimation, especially for systems with smooth dynamics~\cite{suh2022differentiable}.
Unlike model-free RL methods, which rely solely on actual environment interactions, in model-based RL, the actor-critic is trained using imagined trajectories generated by the world model.
For a given horizon length $H$, we generate imagined trajectories by stepping through the world model, starting from the initial states $\{\hat{z}_0, x_0\}$. For $t \in \{1, H\}$,
\begin{equation}
\begin{aligned}
x_t, \; y_t &= \text{SSM}(\hat{z}_{t-1}, a_{t-1}, \hat{c}_{t-1}, x_{t-1}) \\
\hat{z}_t &\sim \; \Dist{p}{\phi}{\cdot}{y_t}  \\
\hat{r}_t &\sim \; \Dist{p}{\phi}{\cdot}{y_t, \hat{z}_t} \\
\hat{c}_t &\sim \; \Dist{p}{\phi}{\cdot}{y_t, \hat{z}_t} \\
a_t &\sim \Dist{\pi}{\theta}{\cdot}{y_t, \hat{z}_t}.
\end{aligned}
\label{eq:imaginationproc}
\end{equation}
Unlike during world model training (see equation~\eqref{eq:wm}), where we have access to the posterior stochastic latent $z_t$, during imagination, we rely solely on the prior stochastic latent $\hat{z}_t$.
We sample the initial states $\{\hat{z}_0, x_0\}$, from the data used during world model training while ensuring sufficient context to generate meaningful predictions, and enough diversity among the initial states to learn a robust value function.

The actor $\pi_{\theta}$ is conditioned on both the deterministic representations $y_t$, and the prior stochastic latent $\hat{z}_t$, as shown in equation~\eqref{eq:imaginationproc}.
The policy follows a Gaussian distribution, with actions sampled as $a_t \sim \Dist{\pi}{\theta}{\cdot}{y_t, \hat{z}_t}$.
In practice, the policy is represented with an MLP outputting the mean and variance of a Gaussian distribution of actions.

The critic, similarly conditioned on $y_t$ and $\hat{z}_t$, as well as $\hat{r}_t$ and $\hat{c}_t$, is trained to predict the bootstrapped $\lambda$-returns~\cite{sutton2018reinforcement}, as proposed in~\cite{Hafner2020Dream}. 
The critic's loss function is given by
\begin{equation}
\begin{aligned}
    \mathcal{L}(\psi) \doteq -\textstyle\sum_{t=1}^T \ln \Dist{p}{\psi}{V^\lambda_t }{x_t}\\
V^\lambda_t \doteq \hat{r}_t + \gamma \hat{c}_t \Big(
  (1 - \lambda) v_t +
  \lambda V^\lambda_{t+1}
\Big),
\end{aligned}
\end{equation}
where $\gamma$ is the discount factor, $v_t \doteq \mathbb{E}[\Dist{v}{\psi}{.}{y_t, \hat{z}_t} ] $ is the predicted value from the critic, and $V^\lambda_T \doteq v_T$.

The actor is trained to maximize the expected returns, which are estimated using a combination of short-horizon imagined trajectories and the value function for long-term value prediction (see Fig.~\ref{fig:imaginationprocess}).
To encourage exploration, an entropy regularization term $-\eta \mathbb{H}[a_t|\hat{z}_t]$ is introduced in the actor's loss function to penalize deterministic policies and to promote more diverse actions for improved exploration,
\begin{equation}
\begin{aligned}
\mathcal{L}(\theta) \doteq
\mathbb{E}_{p_\phi,p_\theta}\Big[
  \textstyle\sum_{t=1}^{H-1} \big(
    -
    V^\lambda_t
    -\eta \mathbb{H}[a_t|\hat{z}_t]
\big)\Big],
\end{aligned}
\end{equation}
Unlike DreamerV3~\cite{hafner2023mastering}, we use the first-order gradients backpropagated through the dynamics model. 
\subsection{Privileged World Models}

In environments with high-dimensional and partially observable inputs, such as the ones encountered in vision-based RL, learning informative latent representations is challenging. 
Additional supervision signals are typically required to help the model learn meaningful representations, which are essential for efficient policy optimization~\cite{lillicrap2015continuous, kalashnikov2018scalable}.

When training policies in simulation, it is possible to take advantage of additional observations that are rarely available during real-world deployments, such information is typically referred to as privileged information and can be used in various ways to boost the training performance in simulation~\cite{pinto2018asymmetric,chen2020learning,yamada2024twist,messikommer2025studentinformed}.
We explore the use of privileged information for training the world model. 
In this setup, the world model has access to privileged information during training, which is not available during real-world deployment. 

In this work, we specifically explore the use of privileged state observations in a vision-based setting. 
During training, the image observation, $o_t$, is encoded into a posterior stochastic state, $z_t$, to be processed by the sequence model (see section~\ref{sec:s5wm}). 
We then replace the decoder from equation~\eqref{eq:wm} with a privileged decoder that reconstructs the privileged state observation, $s_t$, rather than the image observation, 

\begin{equation}
    \text{Privileged decoder:} \quad \hat{s}_t \sim \; p(\cdot|y_t, z_t),
\end{equation}
Reconstructing the state involves predicting $s \in \mathbb{R}^{24}$, which is computationally more tractable than inferring a high dimensional image using an expensive convolutional neural network~(CNN) decoder.
Furthermore, learning to reconstruct state observations, rather than images, provides a stronger signal for training the encoder.
This encourages the latent representation to capture information that is more relevant for policy optimization, rather than optimizing for accurate image reconstruction.

\section{Experiments}
\label{sec:experiments}
We design the experiments to answer the following questions:
\begin{itemize}
    \item Does S5WM accelerate the training of model-based RL?
    \item Does S5WM achieve comparable sample efficiency and task reward to state-of-the-art model-based RL?
    \item What are the advantages of using a privileged world model in partially observable environments?
    \item Does S5WM transfer from simulation to a real-world environment?
\end{itemize}

We validate our approach in a drone racing environment and compare our method against both model-based and model-free baselines.
We then conduct a series of ablations that highlight several design choices of our method.
\subsection{Setup}
\label{sec:setup}

Our setup is consistent across all experiments and utilizes the same quadrotor configuration. 
We first perform training in a simulation environment. 
We then validate our approach in the real world. 
For the environment setup, we use a combination of the Flightmare~\cite{yunlong2020flightmare} and Agilicious~\cite{foehn2022agilicious} software stacks.
All experiments were conducted on the same hardware under uniform conditions. 
\\
\subsubsection{Tasks}
\label{sec:task}
We investigate two tasks, both involving a quadrotor flying in a drone racing environment.

For the first task, termed \textit{state-based Split-S}, the quadrotor flies through the Split-S track and has access to the full-state observations.
This task presents highly complex dynamics, requiring the policy to push the drone's agility to the physical limits of the platform.
Due to the complexity of the track, the task demands both long-term memory and high precision, as the drone must plan ahead to navigate the complex maneuvers at high speeds.

The second task, termed \textit{vision-based Figure-8}, involves flying through the Figure-8 track and involves simpler dynamics, with the drone flying at a slower pace.
However, the challenge in this task lies in learning the dynamics model from high-dimensional image observations, which can be challenging due to the partial observability of the resulting environment.
\\
\begin{figure}
    \centering
    \begin{minipage}{0.51\columnwidth}  %
        \centering
        \includegraphics[width=\linewidth, trim=0 0 0 0, clip]{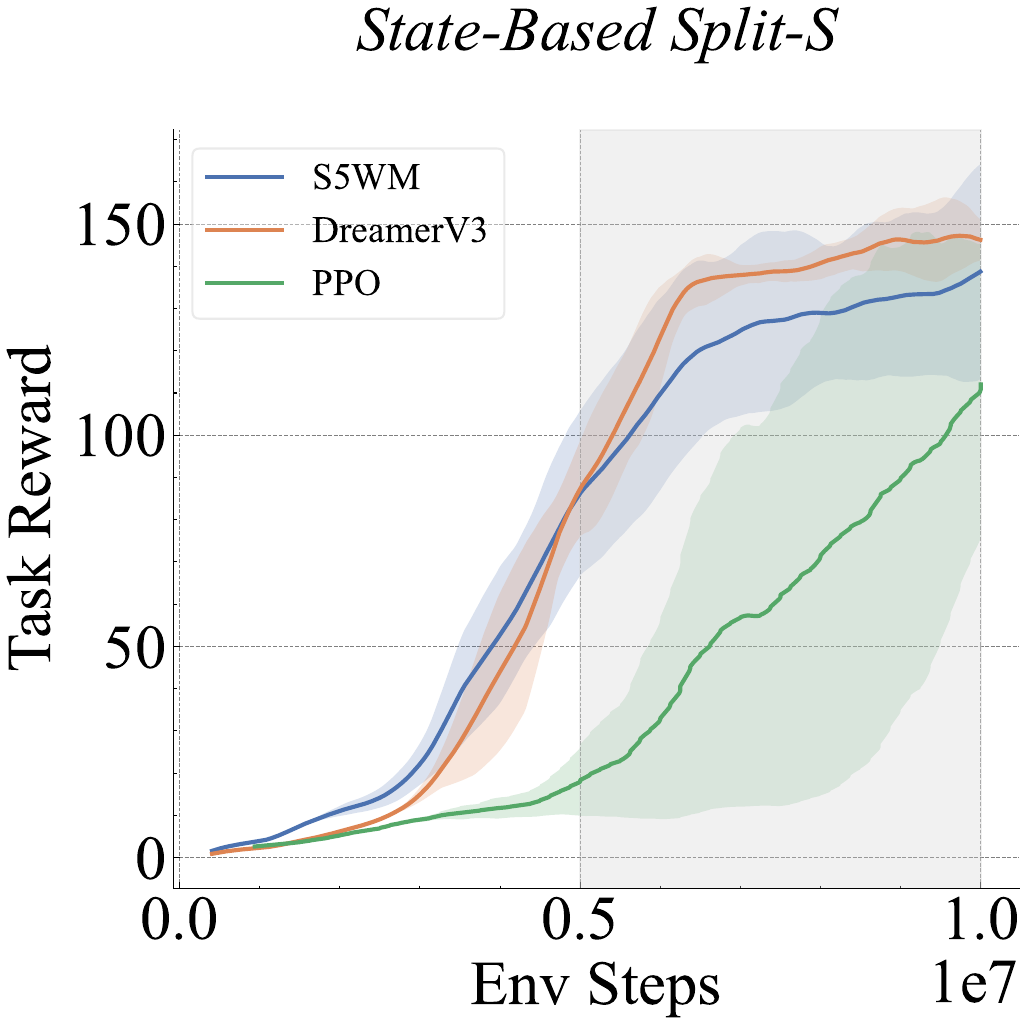}
    \end{minipage}%
    \hfill  %
    \begin{minipage}{0.48\columnwidth}
        \centering
        \includegraphics[width=\linewidth, trim=40 0 0 0, clip]{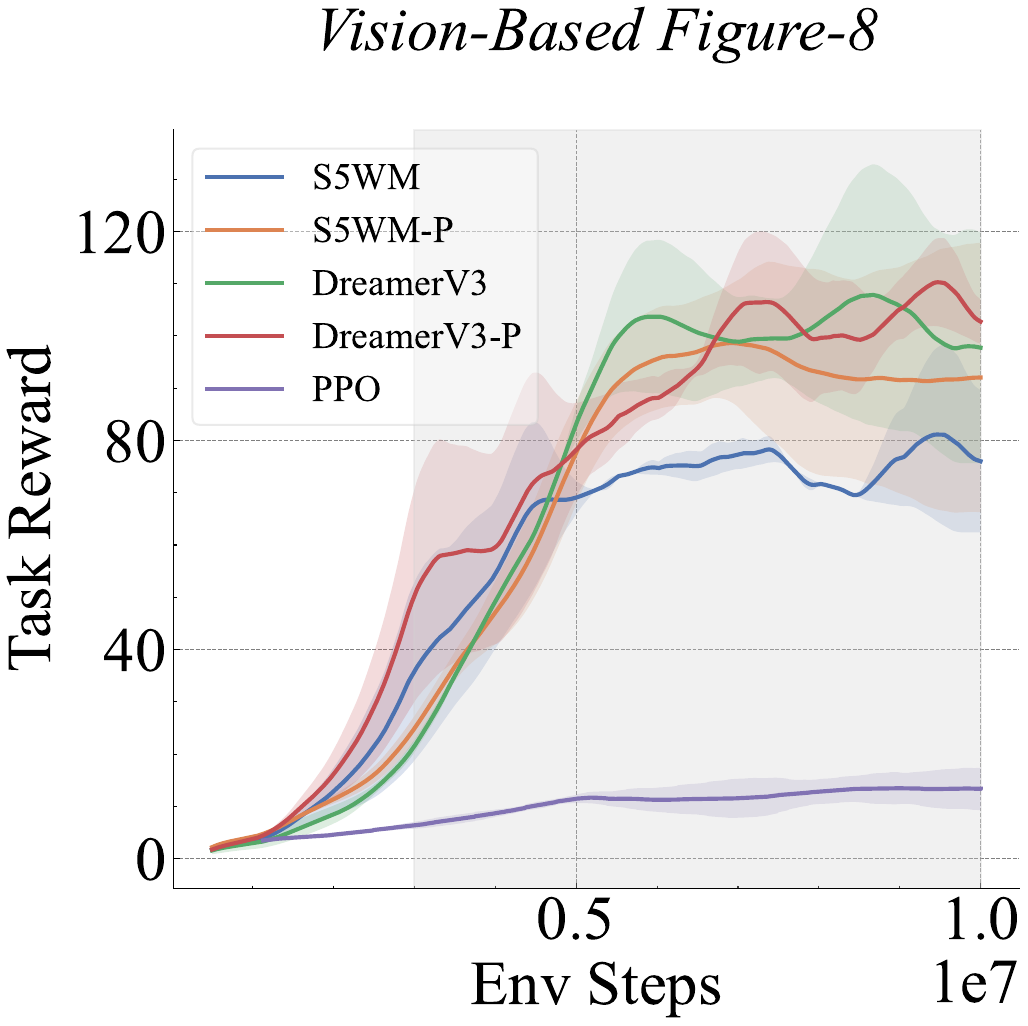}
    \end{minipage}
    \caption{Task reward over the number of environment interactions for S5WM, DreamerV3 and PPO.}
    \label{fig:rewards_state_vision}
\end{figure}

\begin{table}
    \centering
    \caption{Lap Time~(LT) [s] and Success Rate~(SR) [\%] for S5WM (Ours), DreamerV3~\cite{hafner2023mastering} and PPO~\cite{schulman2017ppo} in both simulation, as well as for the real-world experiments. For PPO, the \textit{state-based Split-S} was not available for real-world deployment, while for \textit{vision-based Figure-8}, the asymmetric PPO crashes after the first lap and does not learn to consistently fly through the full track.}
    \label{tab:sim_real_results}
    \setlength{\tabcolsep}{4.5pt} %
    \begin{tabular}{c|c|cc|cc|cc}
        \toprule
        \multirow{2}{*}{\textbf{Task}} & \multirow{2}{*}{\textbf{Env}} 
          & \multicolumn{2}{c}{\textbf{S5WM}} 
          & \multicolumn{2}{c}{\textbf{DreamerV3}} 
          & \multicolumn{2}{c}{\textbf{PPO}} \\
        & & \textbf{LT} & \textbf{SR} 
          & \textbf{LT} & \textbf{SR} 
          & \textbf{LT} & \textbf{SR} \\
        \midrule
        
        \multirow{2}{*}{\textit{State-based Split-S}} 
          & Sim  
            & $4.82$ & 100  
            & $4.79$ & 100  
            & $4.76$ & 100 \\
          & Real  
            & $4.87$ & 100  
            & $4.92$ & 100  
            & - & - \\
        \midrule
        
        \multirow{2}{*}{\textit{Vision-based Figure-8}}  
          & Sim  
            & $3.77$ & 100  
            & $3.87$ & 100  
            & $4.62$ & 44 \\
          & Real  
            & $3.81$ & 100  
            & $3.74$ & 100  
            & - & - \\
        \bottomrule
    \end{tabular}
\end{table}

\subsubsection{Observation Space}
For the \textit{state-based Split-S} task, we define the observation $s = [p, \tilde{R}, v, \omega, i, d, a_{prev}] \in \mathbb{R}^{24}$, where $p \in \mathbb{R}^3$ is the position of the drone, $\tilde{R} \in \mathbb{R}^6$ contains the first two columns of the rotation matrix, $v \in \mathbb{R}^3$ is the linear velocity, and $\omega \in \mathbb{R}^3$ is the angular velocity. The vector $i \in \mathbb{R}^2$ encodes the gate index using sine-cosine encoding to address the periodicity of the track. 
The continuous gate index $i_c$ is defined as 
\begin{equation}
    i_c = i + \frac{2}{1 + \exp(k \cdot d)},
\end{equation}
where $d \in \mathbb{R}$ is the distance to the next gate, and $a_{prev} \in \mathbb{R}^4$ represents the previous action.

For the \textit{vision-based Figure-8} task, the observation space is given by RGB images of size $64 \times 64 \times 3$, rendered from the simulator~\cite{savva2019habitat}.
The raw pixel-level input is then processed through a CNN encoder to extract relevant features.
Similarly to \cite{geles2024demonstrating}, we include the previous 3 actions in the observation to provide historical context. \\
\begin{figure}
    \centering
    \begin{minipage}{0.5\columnwidth}  %
        \centering
        \includegraphics[width=\linewidth, height=4cm, trim=5 0 20 0, clip]{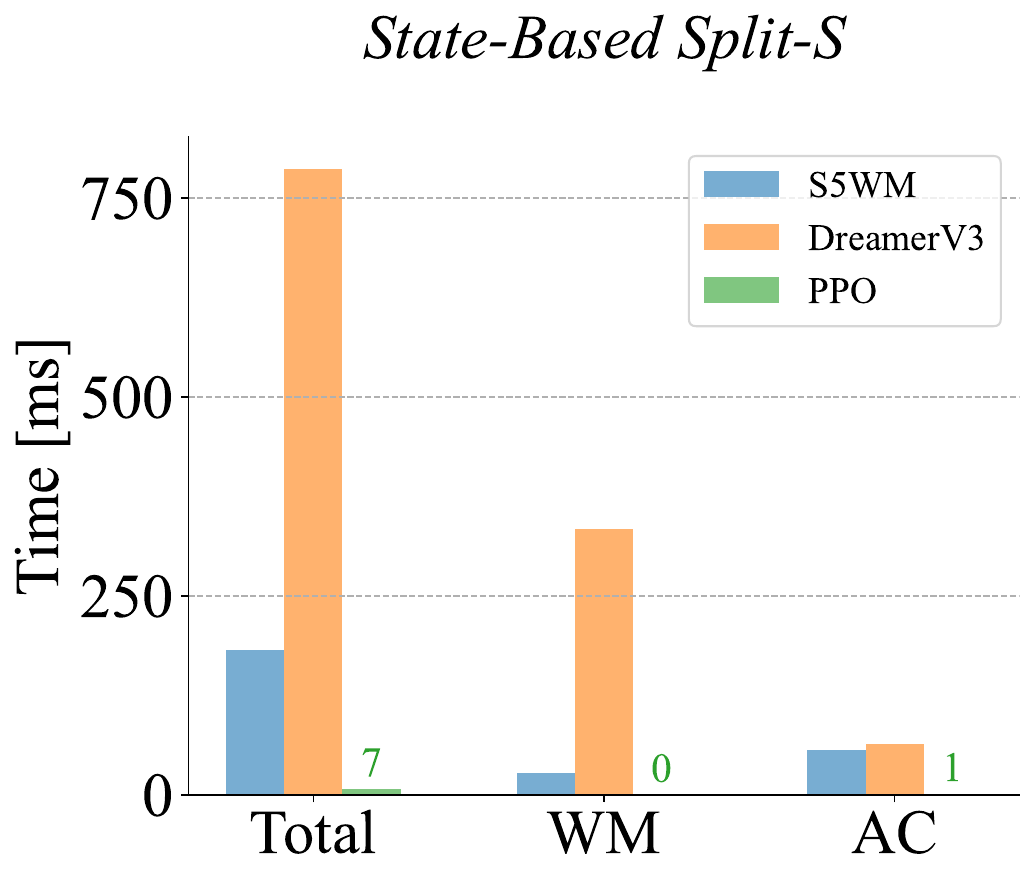}
    \end{minipage}%
    \hfill  %
    \begin{minipage}{0.48\columnwidth}
        \centering
        \includegraphics[width=\linewidth, height=4cm, trim=45 0 20 0, clip]{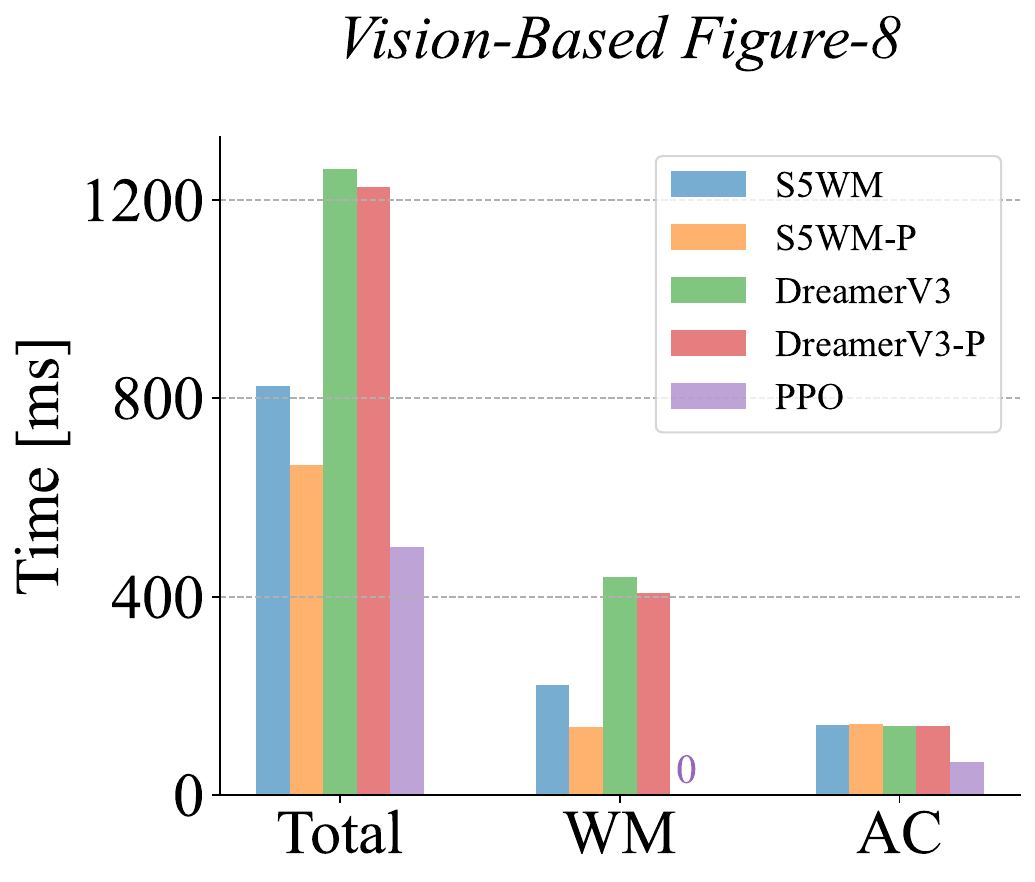}
    \end{minipage}
    \caption{Each training step is divided into: i) training the world model (WM), ii) optimizing the policy (AC), and iii) collecting new data.
    We show the times for each stage, as well as the overall duration per step, averaged over $\text{5}\times\text{10}^5$ steps.}
    \label{fig:timings}
\end{figure}

\subsubsection{Action Space}
\label{sec:actions}
For both tasks, we define $a = [c, \omega_{des}] \in \mathbb{R}^4$, where $c$ is the mass-normalized collective thrust and $\omega_{des} \in \mathbb{R}^3$ represents the desired body rates.
These commands are then processed by a low-level controller, which outputs the desired motor speeds. \\

\subsubsection{Reward}
\begin{figure*}
    \centering
    \begin{minipage}{0.24\textwidth}
        \centering
        \includegraphics[width=\linewidth, trim=0 0 0 0, clip]{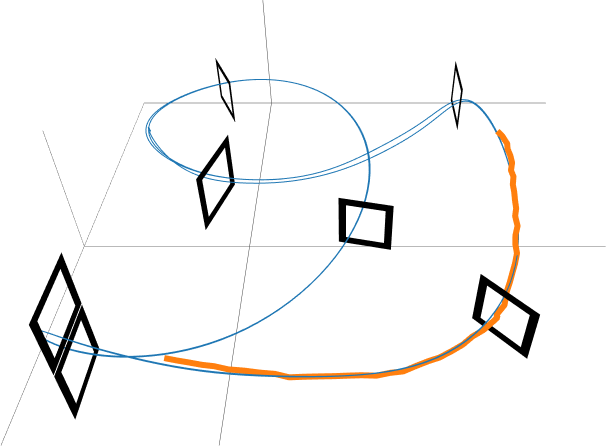}
    \end{minipage}
    \begin{minipage}{0.24\textwidth}
        \centering
        \includegraphics[width=\linewidth, trim=0 0 0 0, clip]{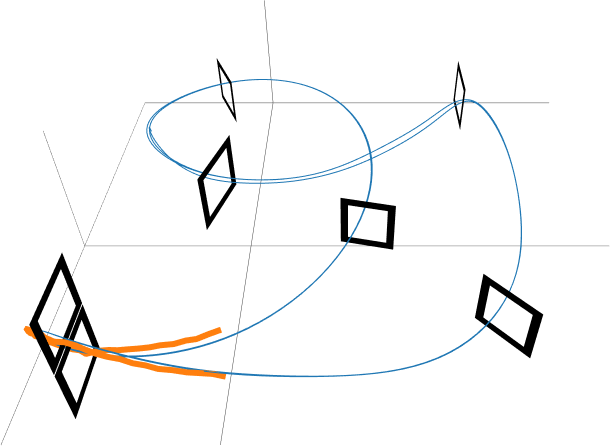}
    \end{minipage}%
    \begin{minipage}{0.24\textwidth}
        \centering
        \includegraphics[width=\linewidth, trim=0 0 0 0, clip]{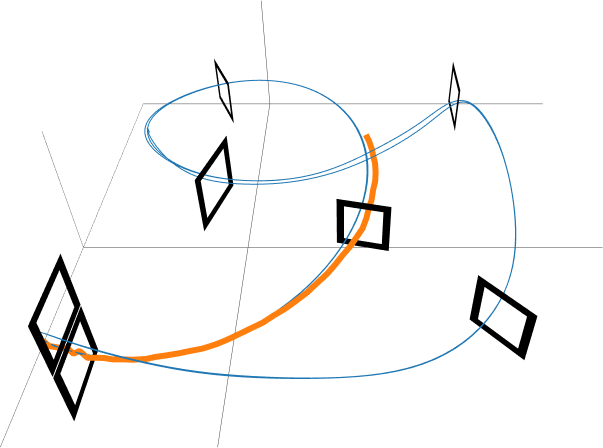}
    \end{minipage}%
    \begin{minipage}{0.24\textwidth}
        \centering
        \includegraphics[width=\linewidth, trim=0 0 0 0, clip]{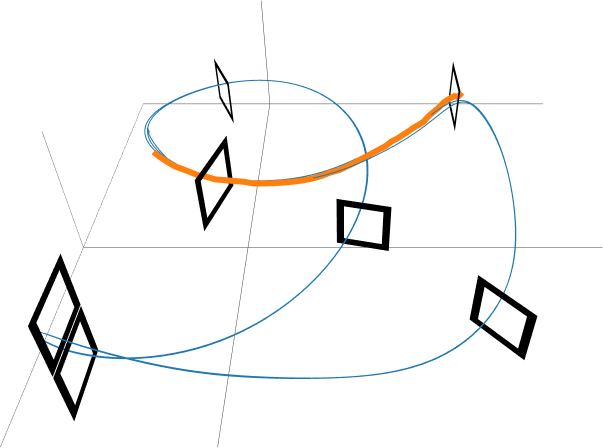}
    \end{minipage}%
    
    \vspace{0.3cm} %
    
    \begin{minipage}{0.24\textwidth}
        \centering
        \includegraphics[width=\linewidth, trim=0 0 20 0, clip]{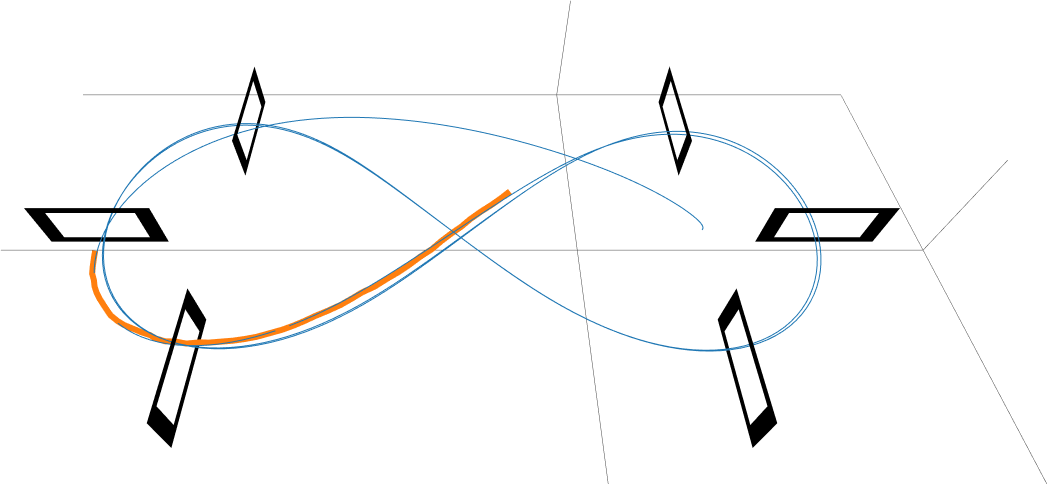}
    \end{minipage}%
    \begin{minipage}{0.24\textwidth}
        \centering
        \includegraphics[width=\linewidth, trim=0 0 20 0, clip]{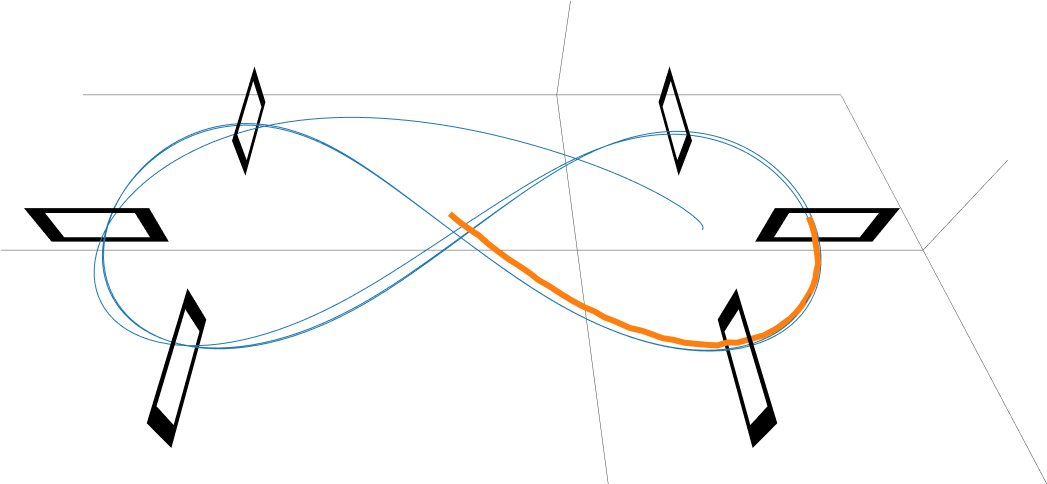}
    \end{minipage}%
    \begin{minipage}{0.24\textwidth}
        \centering
        \includegraphics[width=\linewidth, trim=0 0 20 0, clip]{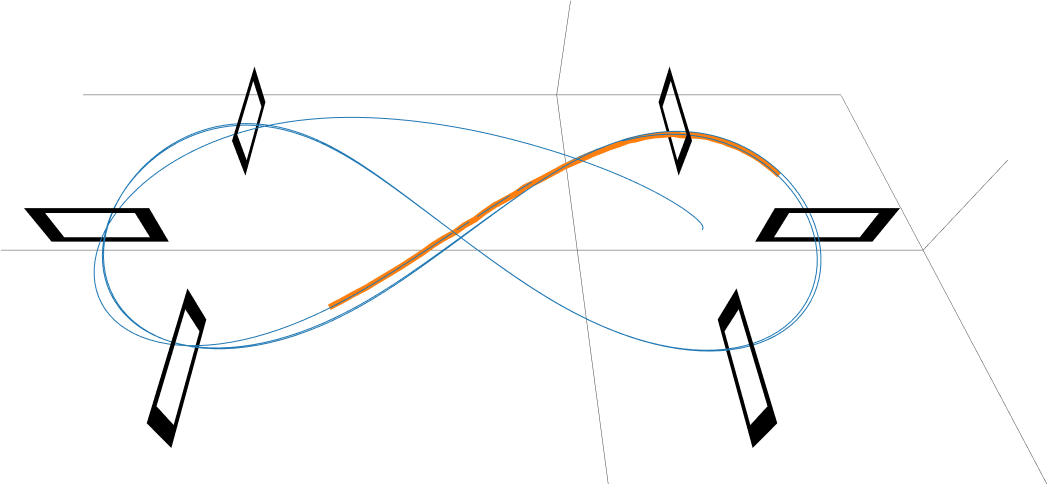}
    \end{minipage}%
    \begin{minipage}{0.24\textwidth}
    \centering
    \includegraphics[width=\linewidth, trim=0 0 20 0, clip]{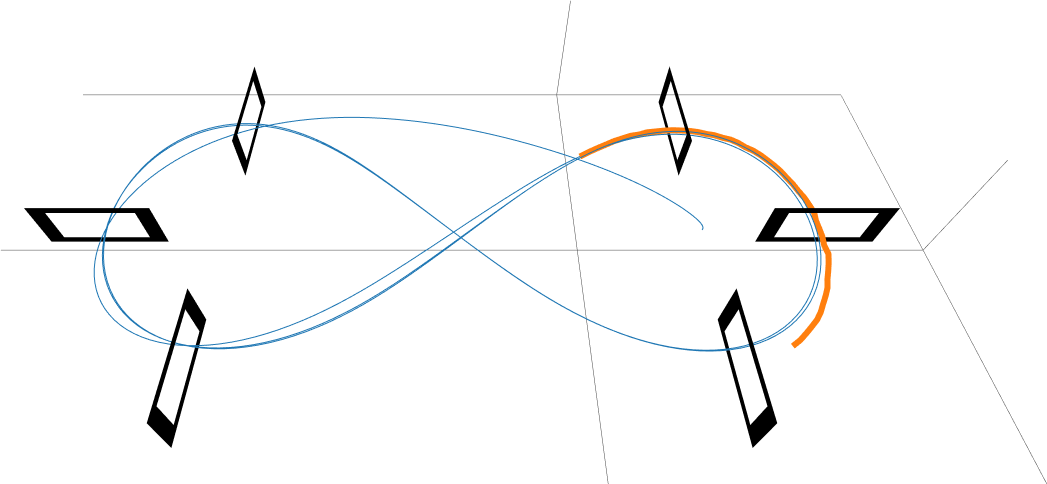}
    \end{minipage}%
    \begin{minipage}{\textwidth}
    \caption{TODO: quality is really bad}
    \end{minipage}%
    \caption{Imagined trajectories for S5WM on the \textit{state-based Split-S} (top) and \textit{vision-based Figure-8} (bottom) tasks over $H=50$ imagination steps, initialized with a context length $C=16$.}
    \label{fig:imaginationtraj}
\end{figure*}
Similar to prior works on drone racing~\cite{romero2024actor, song2023reaching,xing2024multi}, we encode the task using a dense reward function $r_t$,
\begin{equation}
r_t = 
\begin{cases}
r_{\text{crash}} & \text{crashed} \\
r_{\text{pass}} & \text{gate passed} \\
r_{\text{prog}} + r_{\text{omega}} + r_{\text{cmd}} + r_{\Delta\text{cmd}} & \text{otherwise}.
\end{cases}
\end{equation}
where $r_{\text{crash}}$ is a terminal penalty for collisions, $r_{\text{pass}}$ encourages passing a gate, $r_{\text{prog}}$ encourages progress along the track, $r_{\text{omega}}$ penalizes excessive body rates, $r_{\text{cmd}}$ penalizes aggressive actions, and $r_{\Delta\text{cmd}}$ encourages smooth actions. The individual reward components are detailed in the appendix~\ref{sec:rewards}.
\subsection{Baselines}
\label{sec:baselines}
We compare our approach against both model-based and model-free baselines. For the model-based baseline, we use DreamerV3~\cite{hafner2023mastering, romero2024dream}, while for the model-free baseline, we choose PPO~\cite{schulman2017ppo}.
For the vision-based task, we introduce an additional baseline that modifies DreamerV3 to decode privileged state information instead of observation decoding.
We refer to this baseline as "DreamerV3-P".
We also compare against a variant of our method which omits decoding the privileged information and instead decodes raw observations.
Our approach uses the same model architecture and hyperparameters as DreamerV3, with the exception of the world model configuration, where we replace the RSSM with S5WM, as introduced in section~\ref{sec:method}.
We design S5WM to have a comparable number of parameters to RSSM, as detailed in the appendix~\ref{sec:hyperparams}.
We tune the hyperparameters of the baselines to ensure a fair comparison.
For the \textit{state-based Split-S} task, we use standard PPO, and for the \textit{vision-based Figure-8} task, we use PPO with the asymmetric actor-critic architecture~\cite{pinto2018asymmetric}, which provides privileged information to the critic, and employs the same CNN encoder architecture as used in our method and DreamerV3.
We train the CNN encoder jointly with the policy.
For each task, all models share the same observation, action, and reward configurations, as described in Section \ref{sec:setup}.

\subsection{Simulation Results}
We train the policies in a high-fidelity simulator on an A100 GPU.
For all experiments, we simulate 50 environments in parallel and limit the number of interactions with the environment to $\text{10}^{7}$. 
We use a fixed-size replay buffer containing ${10}^{6}$ samples, from which we uniformly sample at each training step.
During evaluation, we rollout the policy over 1000 steps, which corresponds to 20s of flight given a 50Hz control rate.
We report the average task reward obtained per episode.
We evaluate our approach based on performance, accuracy and training efficiency. \\
\begin{figure*}
    \centering
    \begin{minipage}{0.48\textwidth}
        \centering
        \includegraphics[width=\linewidth, trim=0 20 0 20, clip]{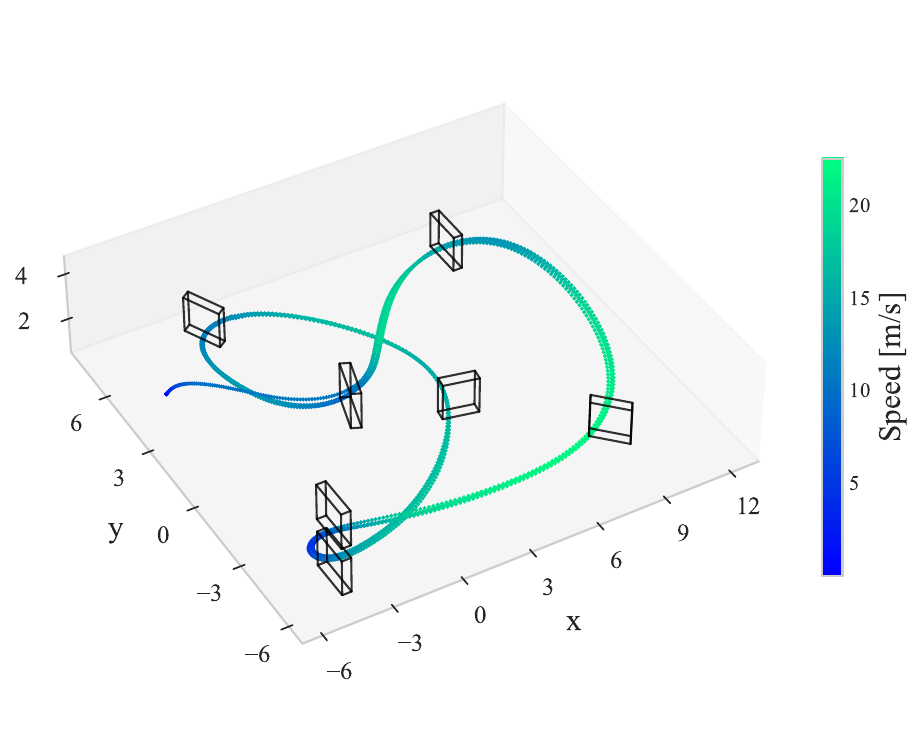}
    \end{minipage}%
    \begin{minipage}{0.48\textwidth}
        \centering
        \includegraphics[width=\linewidth, trim=0 20 0 20, clip]{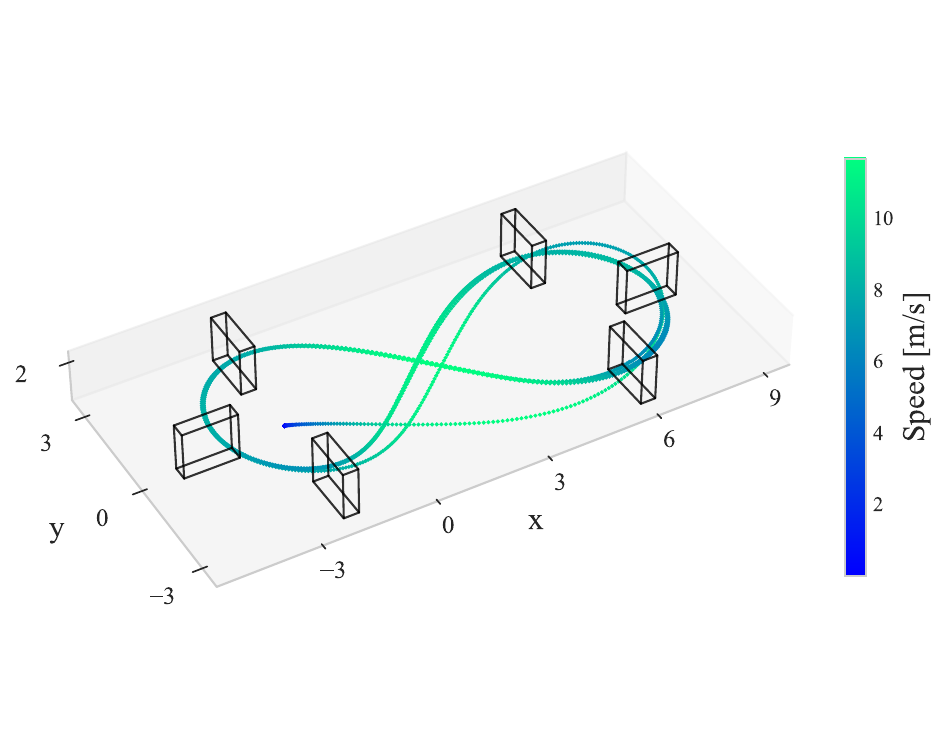}
    \end{minipage}%
    \caption{Real world flight trajectories with S5WM: \textit{State-based Split-S} task (left) and \textit{Vision-based Figure-8} task (right).}
    \label{fig:real_world}
\end{figure*}

\subsubsection{Performance}
In Fig.~\ref{fig:rewards_state_vision}, we evaluate the average task reward over the number of environment interactions for each task.
Overall, we observe a clear advantage of MBRL approaches in terms of sample efficiency.
This result corroborates previous findings from the MBRL literature~\cite{Hafner2020Dream, janner2019trust, yu2020mopo}.
For the \textit{state-based Split-S} task, we find that both S5WM and DreamerV3 achieve similar sample efficiency and converge to the maximum reward, while PPO struggles to yield competitive policies within the limited data budget.

For the \textit{vision-based Figure-8} task, we compare both S5WM and DreamerV3 with and without the privileged world model, as well as the asymmetric PPO.
We find that leveraging privileged information results in higher sample efficiency for both S5WM and DreamerV3.
These benefits are more pronounced for S5WM, where the posterior stochastic latent $z_t$ is inferred directly from the observation.
In contrast, RSSMs also leverage the deterministic state $y_t$ to infer the posterior $z_t$.
These differences are further discussed in section~\ref{sec:ablation_recurrent}.
Overall, our method performs on par with the DreamerV3 variants while being substantially faster to train as we discuss in the next section.
This is evident from the final lap time and success rate achieved by our method compared to the baselines, as shown in Table~\ref{tab:sim_real_results}.

The trajectories obtained after reward convergence are shown in appendix~\ref{sec:appendix_trajectories}.
For both tasks, we find that the most difficult part is completing the track after learning how to fly through the gates.
Additionally, for the \textit{vision-based Figure-8} task, the drone struggles to transition from Gate 3 to Gate 4, as it tends to overshoot and is unable to \textit{see} the next gate.

\subsubsection{Training Efficiency}
We profile S5WM, along with the baselines, to assess the differences in training times.
Each training step is divided into three stages: i) training the world model (observation), ii) optimizing the policy (imagination), and iii) collecting new data by interacting with the environment.
Fig.~\ref{fig:timings} shows the average times for each stage, as well as the overall duration per step.
We find that S5WM outperforms DreamerV3 in terms of overall training speed in both tasks, as the observations in S5WM are parallelized over the sequence length.
In the \textit{state-based Split-S} task, the observation step is up to 10 times faster, leading to an overall speedup of up to 4 times.
In the \textit{vision-based Figure-8} task, the observation step is up to 4 times faster, leading to an overall speedup of up to 2 times.
These benefits are most pronounced when the main computational bottleneck lies in modeling the dynamics, such as in the \textit{state-based Split-S} task.
In tasks that require learning a latent representation from complex visual inputs, such as the \textit{vision-based Figure-8} task, parallelizing the sequence model contributes less to the overall speedup.

\subsubsection{Prediction Accuracy}
We further evaluate the prediction accuracy of the world model.
We provide an initial context length $C=16$ to build a history of hidden states $x_{1:C}$, followed by imagination over the horizon $H$.
During the context period, the world model has access to the true observations from the environment, while during the imagination period, the world model predicts the trajectory without access to the observations.
Fig.~\ref{fig:imaginationtraj} shows the trajectories imagined by S5WM for $H=50$ for different parts of the track, along with the ground truth, which is obtained by interacting with the environment.
A more detailed representation of the imagined trajectories can be found in appendix~\ref{sec:imag_state}.

\subsection{Real World Deployment}

We test our approach in the real world using a high-performance racing drone~\cite{foehn2022agilicious}.
At deployment time, our method runs on an offboard desktop computer equipped with an Intel(R) Core(TM) i9-9880H CPU @ 2.30GHz and Quadro RTX 4000 GPU. 
A Radix FC board equipped with the Betaflight\footnote{\url{https://www.betaflight.com}} firmware is used as the low-level controller, which takes as inputs the desired body rates and collective thrusts. 
An RF bridge is employed to transmit commands to the drone. 
For state estimation, we use a VICON system with 36 cameras that provide the platform with millimeter accuracy measurements of position and orientation at a rate of 400 Hz.
We select the best policy obtained during training and deploy it in the real world.
A video of the real-world deployment of both tasks can be found in the supplementary material.

In addition, Fig.~\ref{fig:real_world} shows the trajectories from the real-world experiments. 
 For the \textit{state-based Split-S} task, the Split-S maneuver at $x=-4.3\unit{m}$ and $y=-5.1\unit{m}$, stands out as a critical test of each approach's characteristics.
This complex maneuver requires the drone to fly through a higher gate and then immediately descend through a second gate located directly below the first one, with both gates sharing the same $x,y$ coordinates. 
This is the most challenging maneuver of the Split-S track, significantly influencing the overall lap time.

For the \textit{vision-based Figure-8} task, we use a hardware-in-the-loop~(HIL) setup, where images are rendered in real-time from the simulator based on state estimation from a motion-caption system estimating the position of the real robot.
For both tasks, our method enables high-speed flight on very smooth trajectories.
\subsection{Ablations}
\label{sec:ablations}
We discuss several key components that we find to be essential to the success of our approach. 
\subsubsection{Horizon Length}
\begin{figure}
    \centering
    \begin{minipage}{0.5\columnwidth}  %
        \centering
        \includegraphics[width=\linewidth, trim=0 0 0 40, clip]{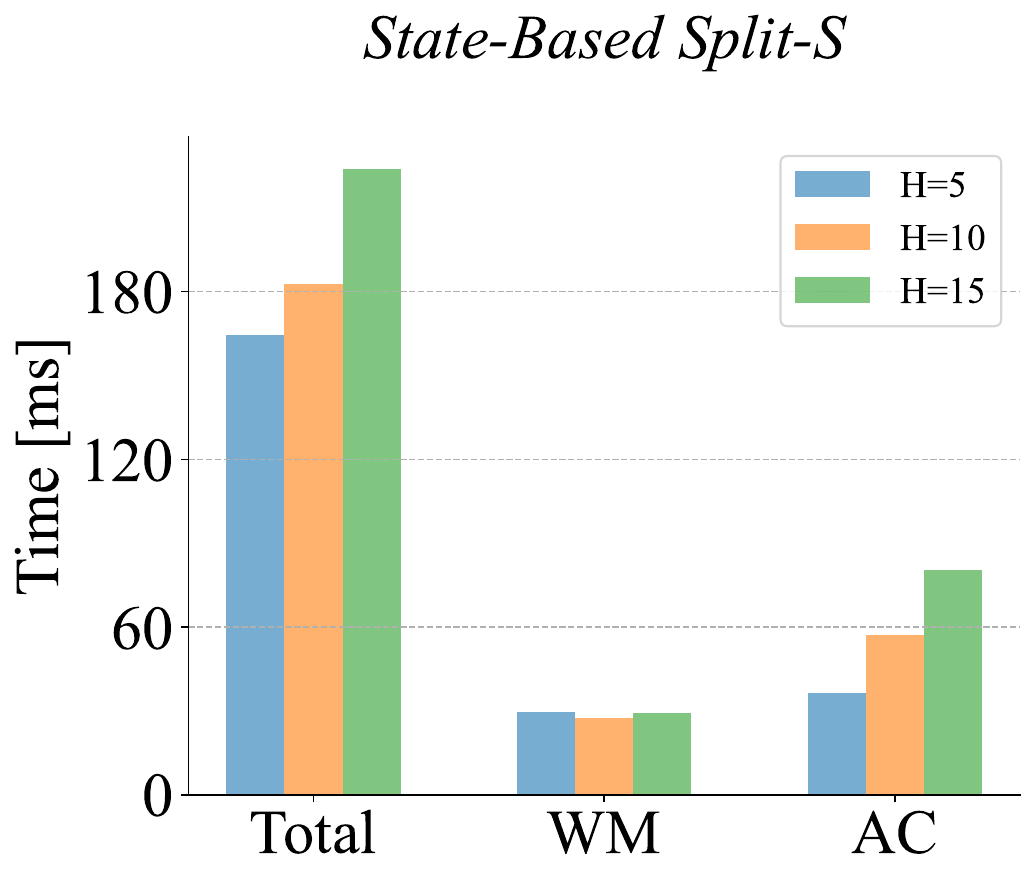}
    \end{minipage}%
    \hfill  %
    \begin{minipage}{0.48\columnwidth}
        \centering
        \includegraphics[width=\linewidth, trim=0 0 0 0, clip]{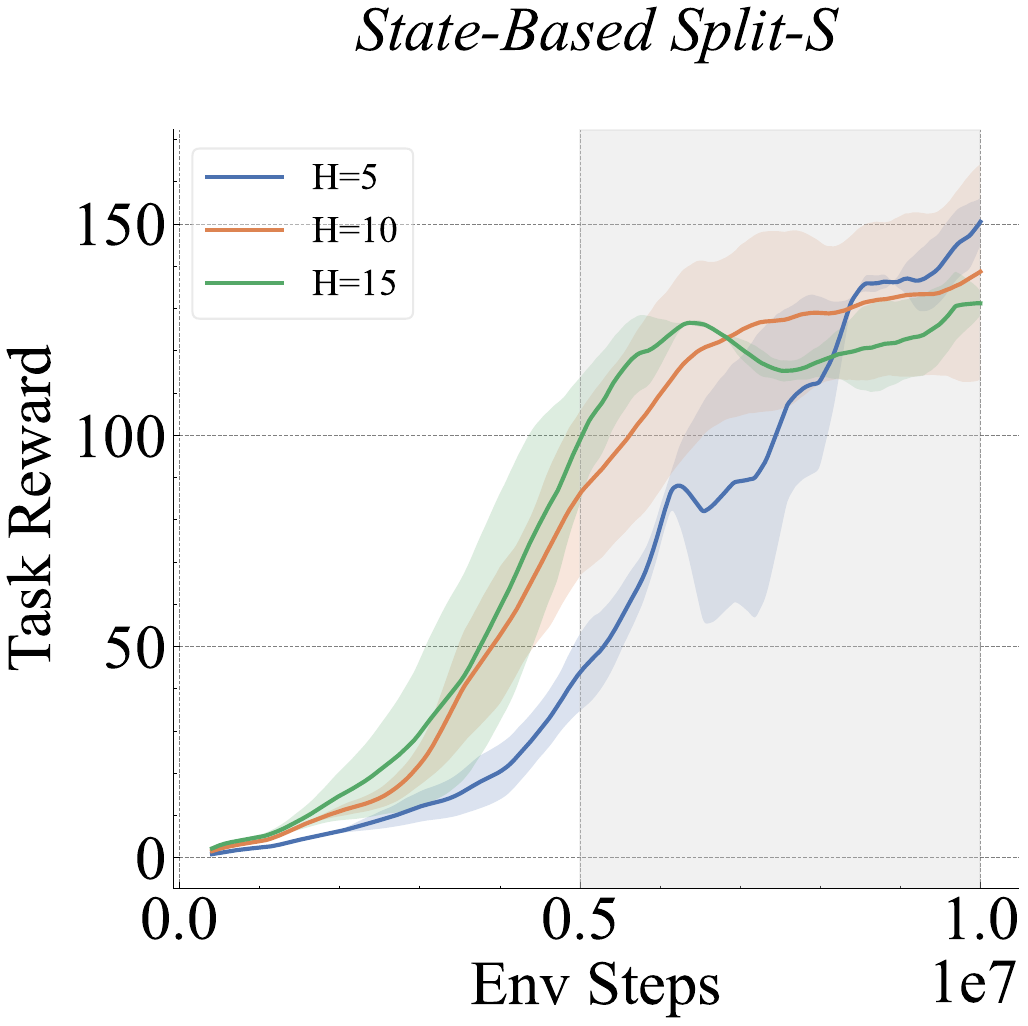}
    \end{minipage}
    \caption{Task reward over the number of environment interactions on S5WM for different imagination horizon lengths $H \in \{5, 10, 15\}$.}
    \label{fig:horizon}
\end{figure}

In model-based RL, the actor-critic uses the world model to plan over a prediction horizon, which can substantially improve sample efficiency compared to model-free RL.
While a longer horizon is typically preferred, prediction errors tend to compound over time and therefore limit the effectiveness of imagination. 
Additionally, imagination steps are computationally expensive and cannot be parallelized, making longer horizons computationally more expensive. 
To investigate the influence of the horizon length on our method, we train the \textit{state-based Split-S} task using different horizon lengths. 
Fig.~\ref{fig:horizon} shows the task reward over the number of environment interactions for $H \in \{5, 10, 15\}$, as well as the time it takes till convergence. 
We find that $H=10$ offers a good balance between sample efficiency and computational complexity, and we therefore use it for all of our experiments.

\subsubsection{Reward Smoothing}
Reward smoothing is essential for enhancing training. 
In model-free RL, it is common to adapt the reward function during training to increase the task difficulty.
While this helps the agent adapt to progressively harder settings, we find that directly altering the reward function in a model-based RL setting can be counterproductive, because the world model needs to \textit{unlearn} the old reward signal and relearn the new one.

\begin{figure}
    \centering
    \begin{minipage}{\columnwidth}
        \centering
        \includegraphics[width=\linewidth, trim=10 10 10 0, clip]{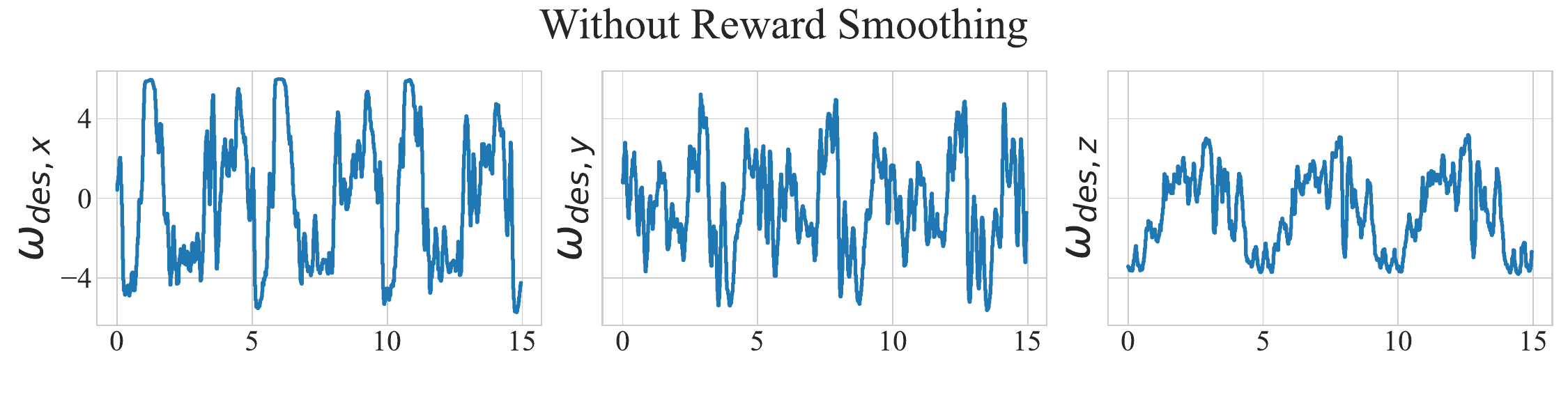}
    \end{minipage}\\[1ex]
    \begin{minipage}{\columnwidth}
        \centering
        \includegraphics[width=\linewidth, trim=10 10 10 0, clip]{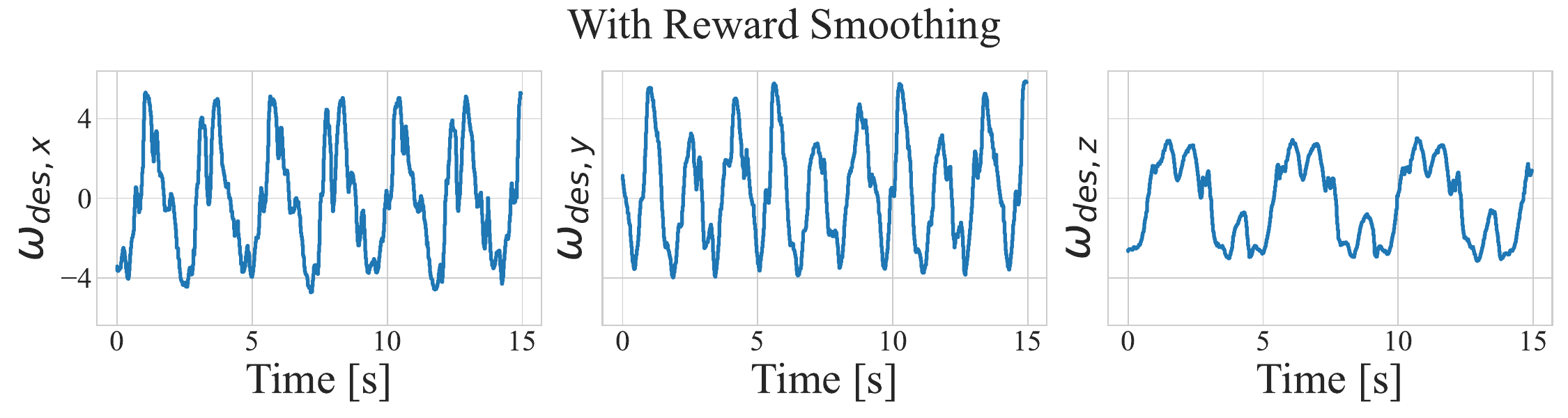}
    \end{minipage}
    \caption{Desired body rates (actions) learned with and without reward smoothing on the \textit{state-based Split-S} task.}
    \label{fig:smooth_actions}
\end{figure}

To address this issue, we propose a smoothing strategy that decouples the total reward into two complementary components,
\begin{equation}
r = r_{\text{nom}} + r_{\text{aug}},
\end{equation}
where $r_{\text{nom}}$ (the nominal reward) is active from the start of training, and $r_{\text{aug}}$ (the smoothing reward) is learned as well from the start but only applied once the agent becomes sufficiently proficient. 
Initially, the value function only receives $r_{\text{nom}}$, and later $r_{\text{aug}}$ is also added.
We use this strategy to encourage smoother actions, which is particularly important for real-world deployment. 
We define
\begin{equation}
\begin{aligned}
r_{\text{nom}}&=r_{\text{prog}}+r_{\text{omega}}+r_{\text{cmd}} \\
r_{\text{aug}}&=r_{\Delta\text{cmd}},
\end{aligned}
\end{equation}
where we follow the definitions from section~\ref{sec:rewards}. 
We apply $r_{\text{aug}}$ halfway through training, after $\text{5}\times\text{10}^{6}$ steps. 
If $r_{\text{aug}}$ is instead applied from the beginning, the agent struggles to learn the task.
By delaying the smoothing reward, the agent first explores to discover a feasible solution, and later refines it to produce smoother actions.
Fig.~\ref{fig:smooth_actions} shows the actions learned with and without reward smoothing.
By introducing this smoothing curriculum, we significantly improve the smoothness of the control commands, preventing motor damage and reducing the sim-to-real gap by ensuring the policy does not learn infeasible command sequences in simulation.

\begin{figure}
    \centering
    \begin{minipage}{0.5\columnwidth}  %
        \centering
        \includegraphics[width=\linewidth, trim=0 0 0 40, clip]{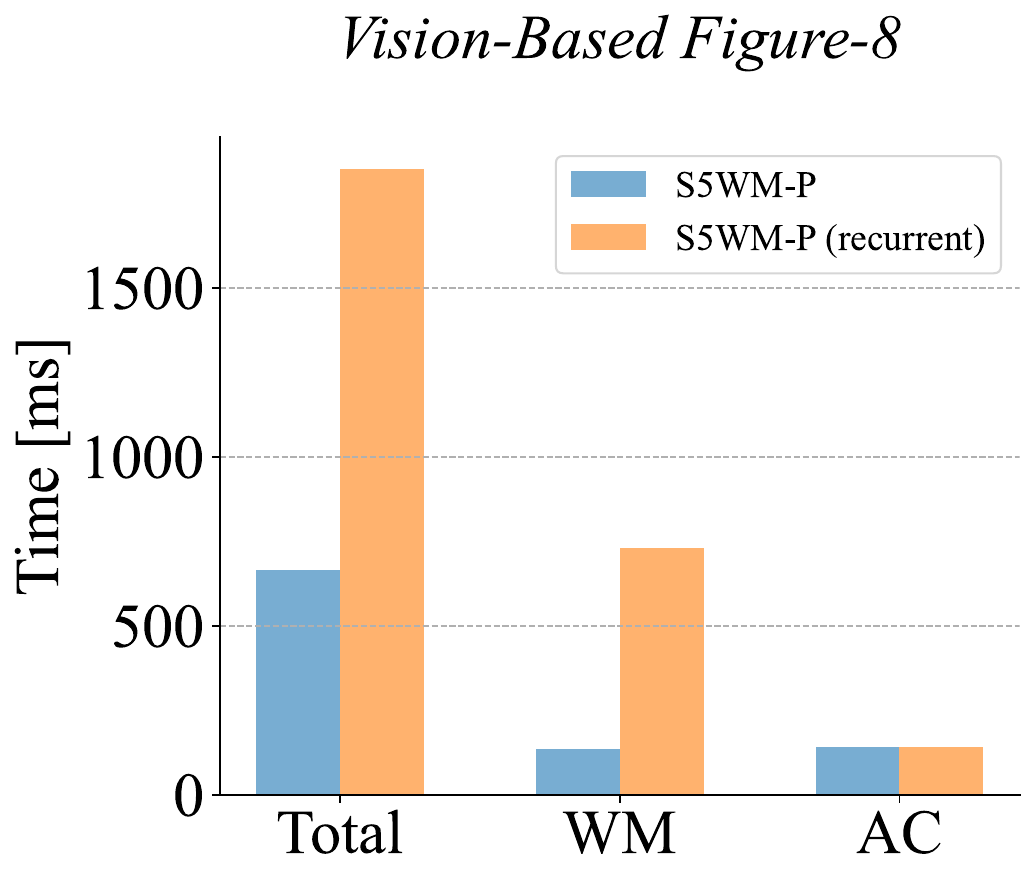}
    \end{minipage}%
    \hfill  %
    \begin{minipage}{0.48\columnwidth}
        \centering
        \includegraphics[width=\linewidth, trim=0 0 0 0, clip]{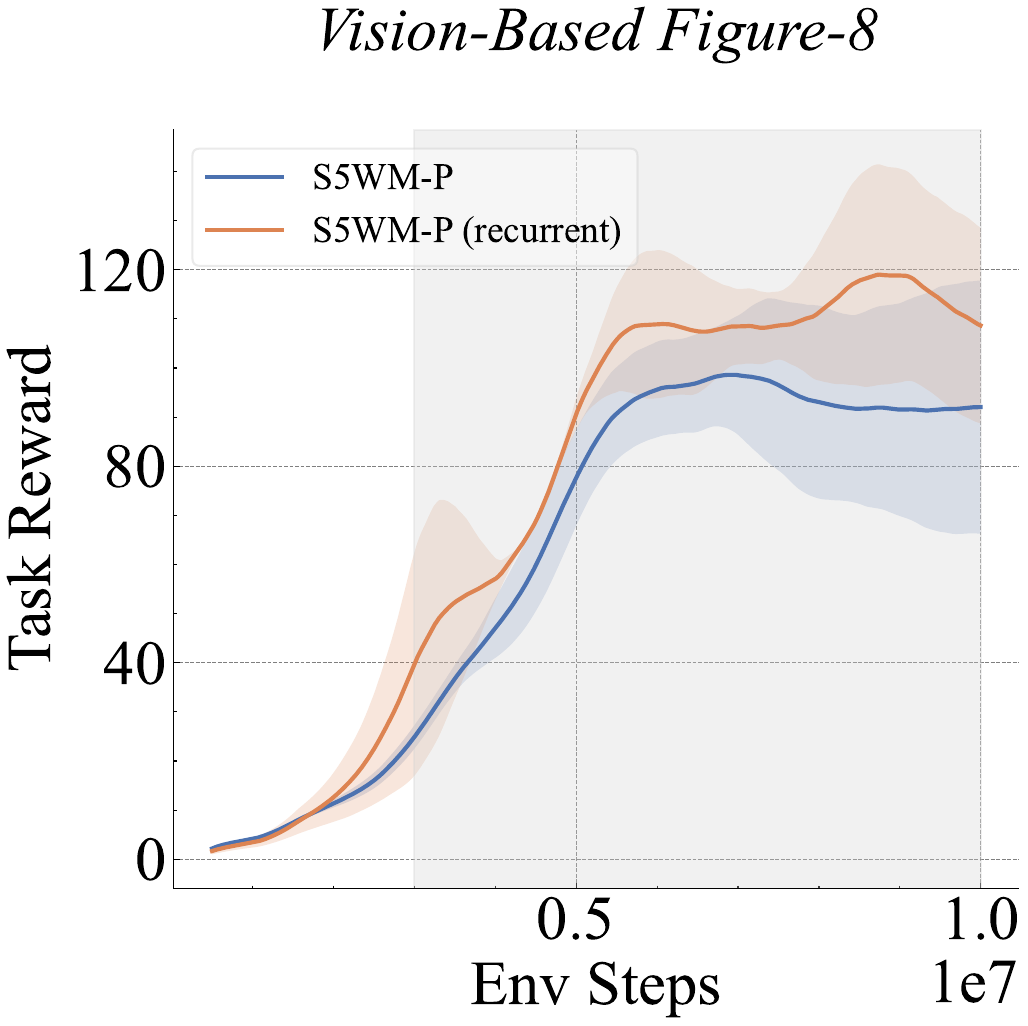}
    \end{minipage}
    \caption{Task reward over the number of environment interactions for parallel and recurrent variants of S5WM.}
    \label{fig:recurrent}
\end{figure}

\subsubsection{Recurrent vs Parallel}
\label{sec:ablation_recurrent}
One main difference between our S5WM architecture (see section~\ref{sec:s5wm}) and RSSMs lies in how the posterior stochastic latent $z_t$ is computed.
In RSSMs, each step computes a prior $\hat{z}_t\sim\;\Dist{p}{\phi}{\cdot}{y_t}$ and a posterior $z_t\sim\;\Dist{q}{\phi}{\cdot}{y_t,o_t}$), both of which include the deterministic representation $y_t$.
Therefore, the posterior receives the same information as the prior plus access to the latest observation $o_t$.
By contrast, since S5WM processes the entire sequence in parallel, the posterior is determined only from the observation, $z_t\sim\;\Dist{q}{\phi}{\cdot}{o_t}$, and does not have access to historical context encoded in $y_t$.
To compensate for this, we append past actions to $o_t$, as explained in section \ref{sec:actions}, providing S5WM with additional temporal context.

We investigate this difference by implementing a \textit{recurrent S5WM} variant where the posterior also depends on $y_t$, i.e., $z_t\sim\;\Dist{q}{\phi}{\cdot}{y_t,o_t}$.
We compare the \textit{recurrent S5WM} to the standard parallel S5WM on the \textit{vision-based Figure-8} task.
Note that the \textit{recurrent S5WM} reintroduces the sequential dependencies of RSSMs, and therefore offers no speed-up in training.
This comparison is included mainly to highlight the importance of historical context in partially observable environments.

\section{Limitations}
While S5WM offers significant advantages in training speed, it also has several shortcomings. 
First, SSMs can be sensitive to hyperparameters, especially compared to RSSMs, which appear more robust across different tasks.
For instance, choosing a suitable architecture of the S5 Block (see appendix~\ref{sec:s5block}), which wraps the S5 layer with nonlinearities and a suitable gating mechanism, is crucial to S5WM's success.
This sensitivity may stem from the rigid, linear nature of the SSMs compared to more flexible architectures like Gated Recurrent Units~(GRUs).
However, a direct comparison of hyperparameter robustness with DreamerV3 would be unfair, given that the latter is fairly mature and has been extensively refined over the years to improve robustness.
A more comprehensive hyperparameter search for S5WM could improve our understanding of its internal workings, though such an analysis would be computationally expensive and remains to be explored in future work.

Finally, the computational benefits of SSMs are most prominent when the bottleneck lies in modeling the dynamics. 
In tasks where the bottleneck shifts to other factors, such as in learning latent representations from high-dimensional visual observations, the advantages of SSMs become less pronounced.

\section{Conclusion}
\label{sec:conclusion}

Despite their notable sample efficiency, model-based reinforcement learning methods can be extremely slow to train, which limits their usability in robotic systems.
This paper introduces S5WM, a state-space world model designed to accelerate training in model-based reinforcement learning. 
We adapt the S5 SSM architecture to handle resetability at the episode boundaries, and use it as a sequence model of the dynamics in our world model. 
Additionally, we introduce asymmetry in the world model by providing privileged state information during training in simulation. 

We compare our approach against model-based and model-free baselines, focusing on training time, performance, and sample efficiency. 
Our method achieves faster training times without compromising performance and sample efficiency. 
Additionally, we test our approach on a drone racing environment and conduct real-world experiments for two distinct tasks involving both state and image observations. 
Our results demonstrate that S5WM performs well in challenging real-world scenarios.
Furthermore, we conduct a series of ablations to highlight the key components of our approach.

While the presented approach shows promising results, we identify several directions for future work. 
For instance, one valuable area of research would be to speed up the actor-critic training in MBRL.
Additionally, our proposed approach is not as robust to hyperparameter changes as more mature MBRL methods such as Dreamer.
Future work could investigate modifications to our method to improve its robustness to hyperparameter choice.
Furthermore, it would be interesting to apply and evaluate our method on different robotic domains such as locomotion and manipulation.

In summary, our method offers a promising framework for accelerating model-based reinforcement learning, combining high sample efficiency with faster training times, making it more feasible for real-world robotic deployment.

\section*{Acknowledgments}
This work was supported by the European Union’s Horizon Europe Research and Innovation Programme under grant agreement No. 101120732 (AUTOASSESS) and the European Research Council (ERC) under grant agreement No. 864042 (AGILEFLIGHT).

\bibliographystyle{unsrtnat}
\bibliography{references}
\clearpage
\newpage
\appendix
\section{Appendix}
\label{sec:appendix}
\subsection{Hyperparameters}
The RSSM is configured as in DreamerV3~\cite{hafner2023mastering}, whereas the parameters related to the world model have been adjusted for the S5WM.
\label{sec:hyperparams}
\begin{table}[h]
\centering
\begin{tabular}{lcc}
 & S5WM & RSSM \\ \toprule
\textbf{General} & & \\ %
Steps & $10^7$ & $10^7$ \\
Optimizer & Adam & Adam \\
Activation & SiLU & SiLU \\
MLP units & 512 & 512 \\
MLP layers & 2 & 2 \\
Batch Size & 32 & 16 \\
Batch Length & 128 & 128 \\
Dataset Size & $10^6$ & $10^6$ \\
Num. Environments & 50 & 50 \\
Discount Factor $\gamma$ & 0.997 & 0.997 \\
Dynamic Factor $\beta_{dyn}$ & 0.5 & 0.5 \\
Representation Factor $\beta_{rep}$ & 0.1 & 0.1 \\ \midrule 
\textbf{World Model} & & \\ %
Learning Rate & $5$ x $10^{-4}$ & $1$ x $10^{-4}$ \\
Deterministic State $y$ & 512 & 512 \\
Stochastic State $z$ & 32x32 & 32x32 \\
Imagination Horizon $H$ & 10 & 10 \\
Imagination Gradient & Dynamics & Dynamics \\
Context Length & 32 & 32 \\ \midrule
\textbf{State Space Model} & & \\ %
Hidden State $x$ & 256 & - \\
Activation & Half GLU & - \\
Num. Layers & 1 & - \\
Degree & 1 & - \\
Learning Rate & $5$ x $10^{-4}$ & - \\ \midrule
\textbf{Actor} & & \\ %
Activation & Tanh & Tanh \\
Distribution & Gaussian & Gaussian \\
Entropy Coefficient $\eta$ & $2$ x $10^{-3}$ & $2$ x $10^{-3}$ \\
Learning Rate & $3$ x $10^{-5}$ & $3$ x $10^{-5}$ \\ \midrule
\textbf{Critic} & & \\ %
Activation & Tanh & Tanh \\
Distribution & symlog\_disc & symlog\_disc \\
Learning Rate & $3$ x $10^{-5}$ & $3$ x $10^{-5}$ \\ \bottomrule \\
\end{tabular}
\caption{Hyperparameters for the \textit{state-based Split-S} task.}
\label{tab:state_params}
\end{table}

\begin{table}[h]
\centering
\begin{tabular}{lcc}
 & S5WM (x$10^6$) & RSSM (x$10^6$) \\ \midrule
Encoder & 0.28 & 0.28 \\
Dynamics & 3.55 & 3.94 \\
Reward & 1.44 & 1.44 \\
Continuity & 1.05 & 1.05 \\
Decoder & 1.06 & 1.06 \\
Actor & 1.05 & 1.05 \\
Critic & 1.18 & 1.18 \\
\midrule
\textbf{Total} & 8.62 & 8.94 \\
\bottomrule
\end{tabular}
\caption{Number of parameters by network component for the \textit{state-based Split-S} task.}
\label{tab:model_params_state}
\end{table}

\begin{table}[h]
\centering
\begin{tabular}{lcc}
 & S5WM & RSSM \\ \toprule
\textbf{General} & & \\ %
Steps & $10^7$ & $10^7$ \\
Optimizer & Adam & Adam \\
Activation & SiLU & SiLU \\
MLP units & 768 & 768 \\
MLP layers & 4 & 4 \\
Batch Size & 32 & 16 \\
Batch Length & 128 & 128 \\
Dataset Size & $10^6$ & $10^6$ \\
Num. Environments & 50 & 50 \\
Discount Factor $\gamma$ & 0.997 & 0.997 \\
Dynamic Factor $\beta_{dyn}$ & 0.5 & 0.5 \\
Representation Factor $\beta_{rep}$ & 0.1 & 0.1 \\ \midrule 
\textbf{World Model} & & \\ %
Learning Rate & $10^{-3}$ & $10^{-4}$ \\
Deterministic State $y$ & 1024 & 1024 \\
Stochastic State $z$ & 32x32 & 32x32 \\
Imagination Horizon $H$ & 10 & 10 \\
Imagination Gradient & Dynamics & Dynamics \\
Context Length & 32 & 32 \\ \midrule
\textbf{State Space Model} & & \\ %
Hidden State $x$ & 512 & - \\
Activation & Half GLU & - \\
Num. Layers & 2 & - \\
Degree & 1 & - \\
Learning Rate & $5$ x $10^{-4}$ & - \\ \midrule
\textbf{Actor} & & \\ %
Activation & Tanh & Tanh \\
Distribution & Gaussian & Gaussian \\
Entropy Coefficient $\eta$ & $2$ x $10^{-3}$ & $2$ x $10^{-3}$ \\
Learning Rate & $3$ x $10^{-5}$ & $3$ x $10^{-5}$ \\ \midrule
\textbf{Critic} & & \\ %
Activation & Tanh & Tanh \\
Distribution & symlog\_disc & symlog\_disc \\
Learning Rate & $3$ x $10^{-5}$ & $3$ x $10^{-5}$ \\ \bottomrule \\
\end{tabular}
\caption{Hyperparameters for the \textit{vision-based Figure-8} task.}
\label{tab:vision_params}
\end{table}

\begin{table}[h]
\centering
\begin{tabular}{lcc}
 & S5WM (x$10^6$) & RSSM (x$10^6$) \\ \midrule
Encoder & 0.69 & 0.69 \\
Dynamics & 13.05 & 12.60 \\
Reward & 3.94 & 3.94 \\
Continuity & 3.35 & 3.35 \\
Decoder & 3.37 & 3.37 \\
Actor & 3.35 & 3.35 \\
Critic & 3.54 & 3.54 \\
\midrule
\textbf{Total} & 31.29 & 30.84 \\
\bottomrule
\end{tabular}
\caption{Number of parameters by network component for the \textit{vision-based Figure-8} task.}
\label{tab:model_params_vision}
\end{table}

\newpage

\subsection{Imagination Prediction}
\label{sec:imag_state}
\begin{figure}[H]
    \centering
    \begin{minipage}{\columnwidth}
        \centering
        \includegraphics[width=\linewidth, trim=0 0 0 0, clip]{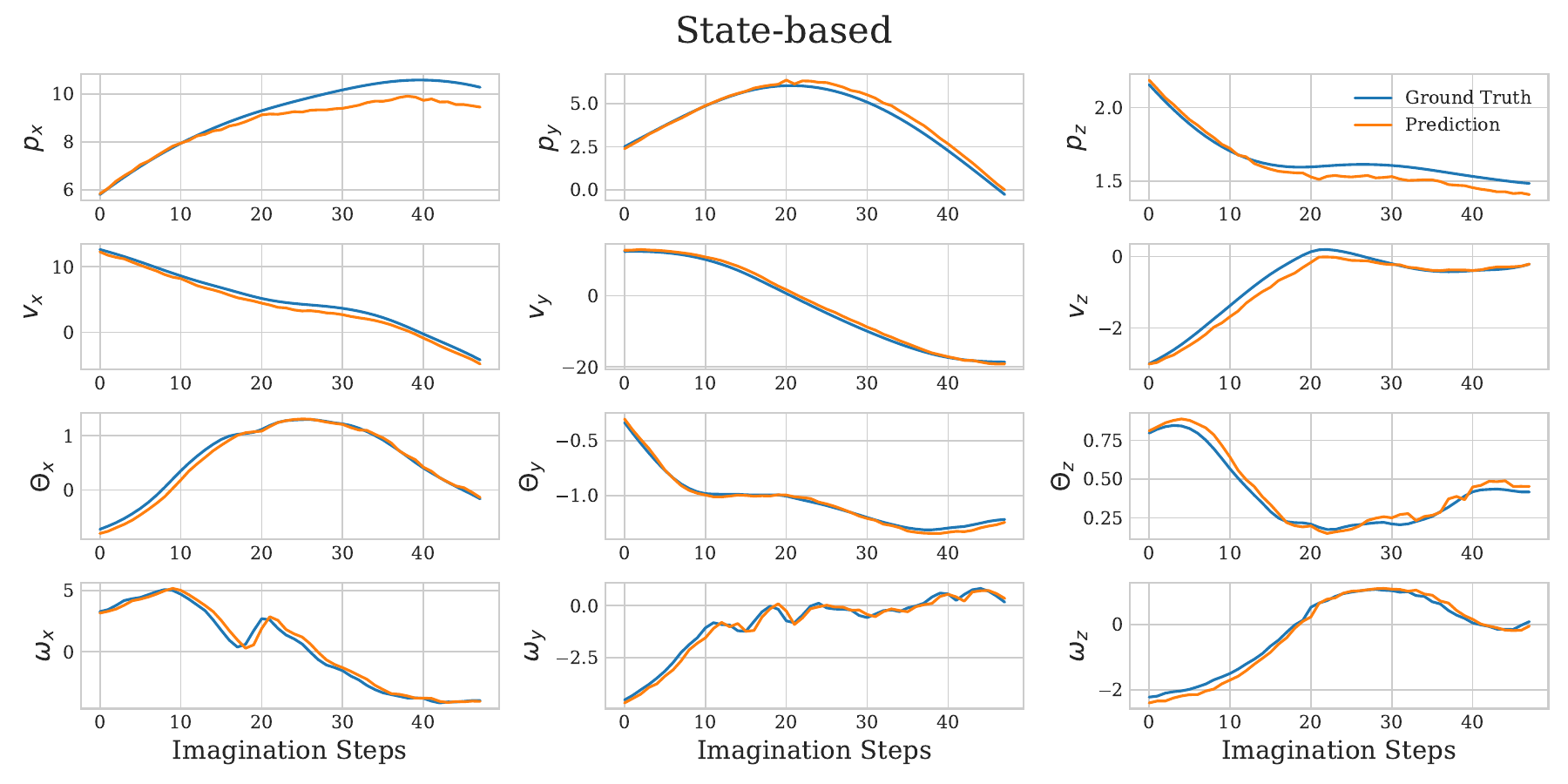}
    \end{minipage}\\[1ex]
    \begin{minipage}{\columnwidth}
        \centering
        \includegraphics[width=\linewidth, trim=0 0 0 0, clip]{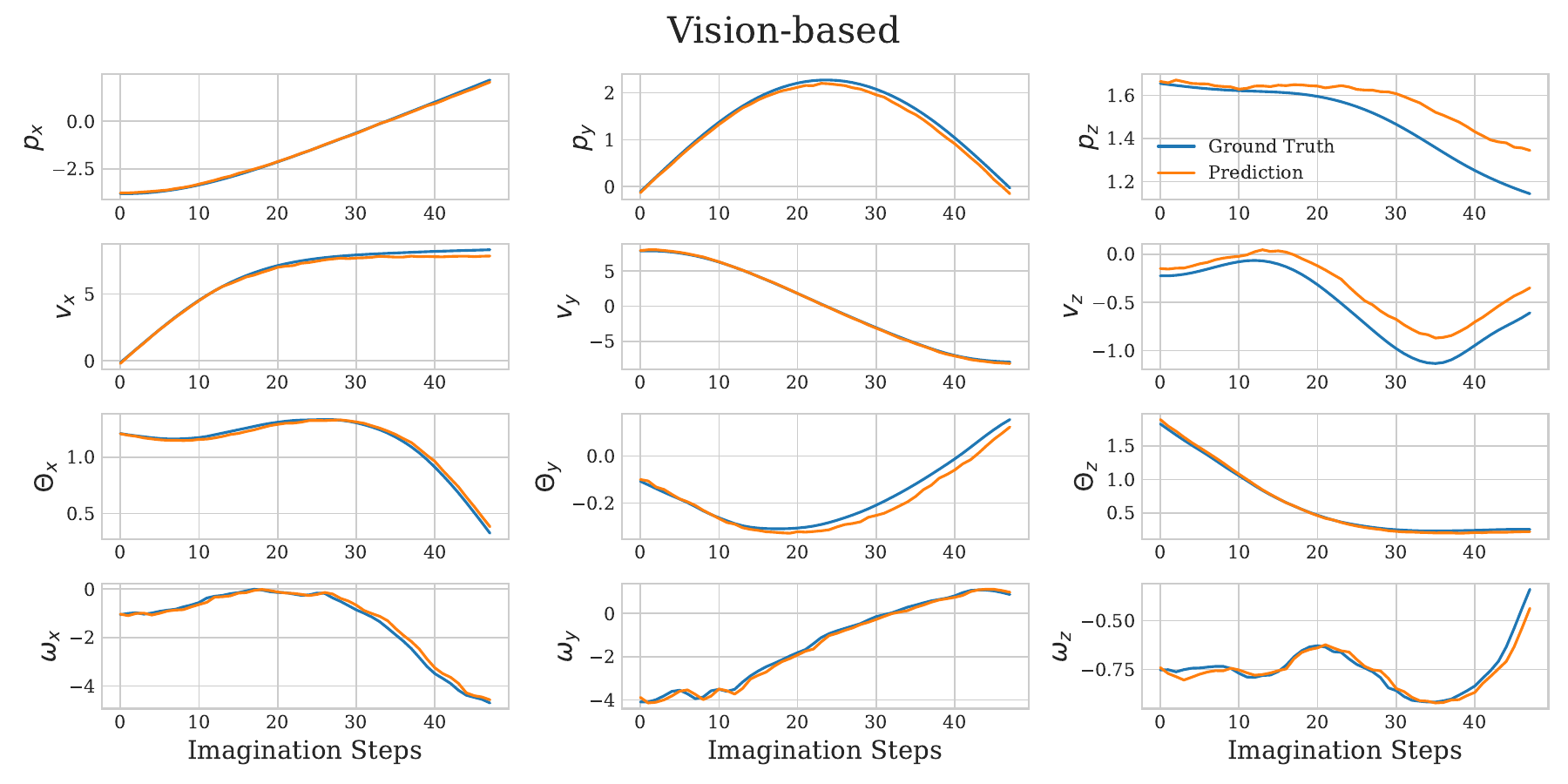}
    \end{minipage}
    \caption{Imagined state for S5WM on \textit{state-based Split-S} and \textit{vision-based Figure-8} tasks over $H=50$ steps.}
    \label{fig:imag_state}
\end{figure}

\subsection{Reward components}
\label{sec:rewards}
\begin{enumerate}\renewcommand{\labelenumi}{\roman{enumi}.} 
    \item \textit{Progress Reward}: Encourages the drone to progress along the track as fast as possible,
    \begin{equation}
    r_{\text{prog}} = \lambda_1 (d_{t-1} - d_t),
    \end{equation}
    where $d_t$ is the distance from the position at time $t$ to the next gate, and $\lambda_1=1$.

    \item \textit{Omega Penalty}: Penalizes large angular velocities,
    \begin{equation}
    r_{\text{omega}} = \lambda_2 \left(\frac{\| \omega \|}{\| \omega_{\text{max}} \|}\right)^2,
    \end{equation}
    where $\omega$ is the angular velocity, $\omega_{\text{max}}$ is the maximum angular velocity, and $\lambda_2=-0.05$.

    \item \textit{Command Omega Penalty}: Penalizes large desired angular velocity values (actions),
    \begin{equation}
    r_{\text{cmd}} = \lambda_3 \left(\frac{\| \omega_{\text{des}} \|}{\| \omega_{\text{max}} \|}\right)^2,
    \end{equation}
    where $\omega_{des}$ is the desired angular velocity command, $\omega_{\text{max}}$ is the maximum angular velocity, and $\lambda_3=-0.025$.

    \item \textit{Command Difference Penalty}: Penalizes abrupt changes between consecutive actions to encourage smooth signals,
    \begin{equation}
    r_{\Delta\text{cmd}} = \lambda_4 \left(\frac{\| \Delta a_t \|}{\|a_{\text{max}}\|} \right)^2,
    \end{equation}
    where $\Delta a_t = a_t - a_{t-1}$, $a_t \in \mathbb{R}^4$ is the control action at time $t$, $a_{\text{max}}$ is the maximum control action, and $\lambda_4=-0.05$.

    \item \textit{Crash Penalty}: Terminal reward that penalizes collisions with the gates,
    \begin{equation}
    r_{\text{crash}} = -4.
    \end{equation}

    \item \textit{Passing Reward}: Encourages the drone to pass through the gate successfully,
    \begin{equation}
    r_{\text{pass}} = 10.
    \end{equation}

\end{enumerate}

\subsection{S5 Sequence Layer}
\label{sec:s5block}
\begin{figure}[h]
    \centering
    \includegraphics[width=0.8\columnwidth, trim=0 0 0 0, clip]{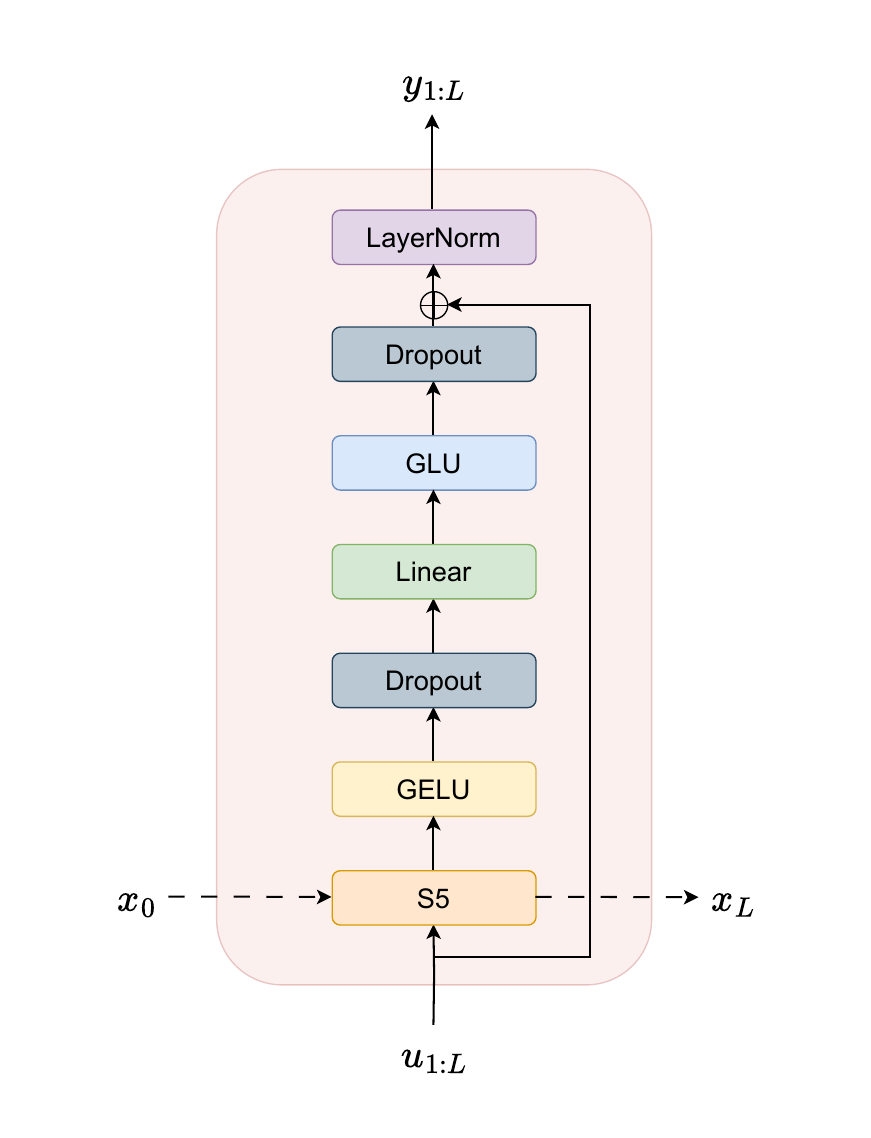}
    \caption{Schematic representation of the S5 sequence layer, composed of the resettable S5 SSM described in~\ref{sec:s5wm}, nonlinearities, and gating operations.}
    \label{fig:reconstruction}
\end{figure}
For the Gated Linear Unit~(GLU), we use a specific gating function, defined as \textit{half GLU}, which maps its input $u$ to the output $y$ as follows,
\begin{align}
x_1 = GELU(u) \\
y = u \cdot \sigma(W^T x_1)
\end{align}
where $W$ represents the weights of a linear layer, $GELU$ is a standard Gaussian Error Linear Unit, and $\sigma$ is the sigmoid function.

\newpage

\subsection{Trajectories}
\label{sec:appendix_trajectories}
\begin{figure}[htp]
    \centering
    \begin{minipage}{0.3\textwidth}
        \centering
        \includegraphics[width=\linewidth, trim=30 10 60 0, clip]{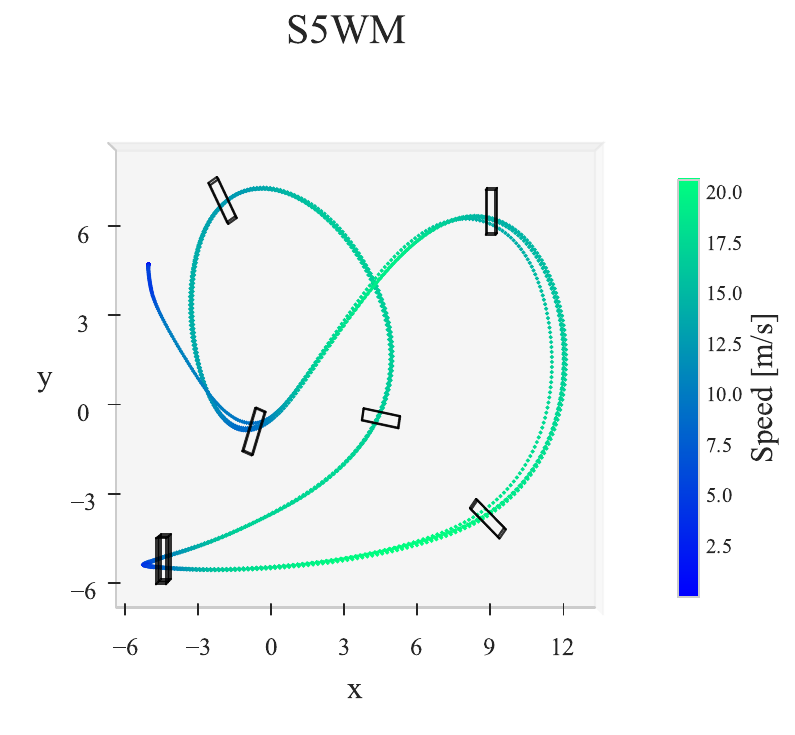}
    \end{minipage}%
    \begin{minipage}{0.3\textwidth}
        \centering
        \includegraphics[width=\linewidth, trim=30 10 60 0, clip]{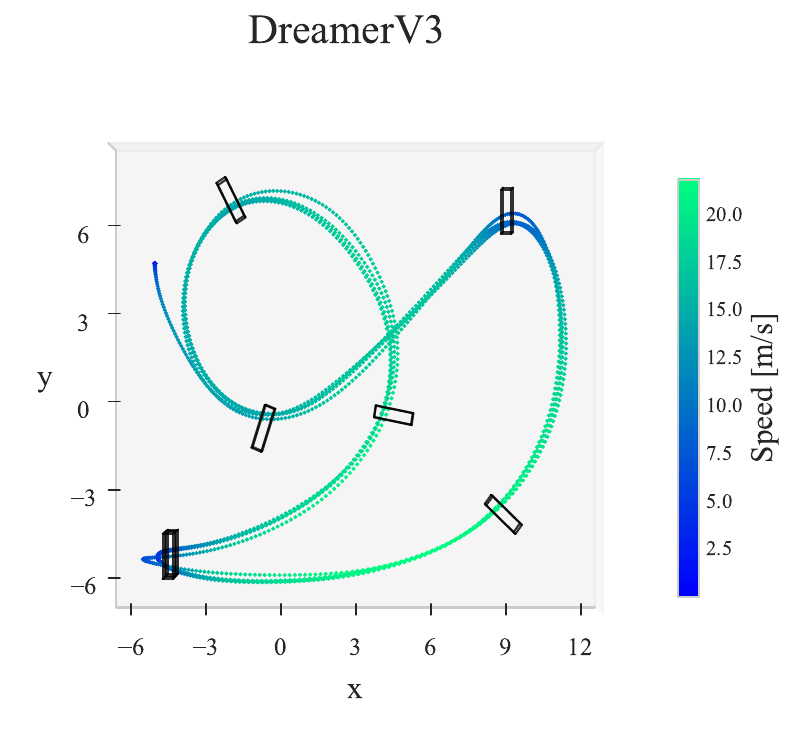}
    \end{minipage}%
    \begin{minipage}{0.35\textwidth}
        \centering
        \includegraphics[width=\linewidth, trim=30 10 0 0, clip]{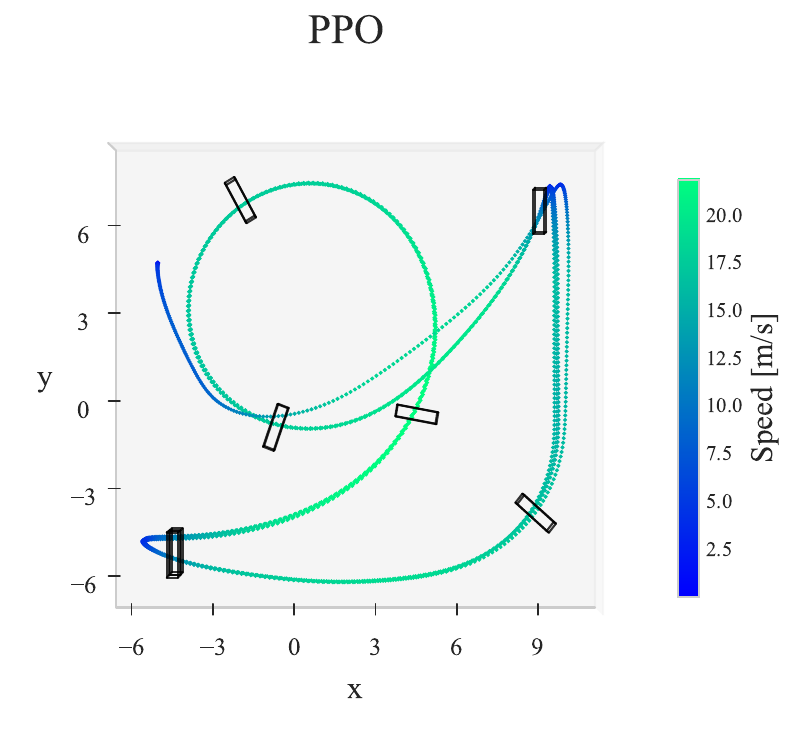}
    \end{minipage}%
    \label{fig:real_splits}
\end{figure}
\begin{figure}[h]
    \begin{minipage}{0.95\textwidth}
    \caption{\textit{State-based Split-S task:} Comparison for S5WM, with baselines: DreamerV3 and PPO.}
    \end{minipage}
\end{figure}
\begin{figure}[htp]
    \centering
    \begin{minipage}{0.3\textwidth}
        \centering
        \includegraphics[width=\linewidth, trim=20 10 60 0, clip]{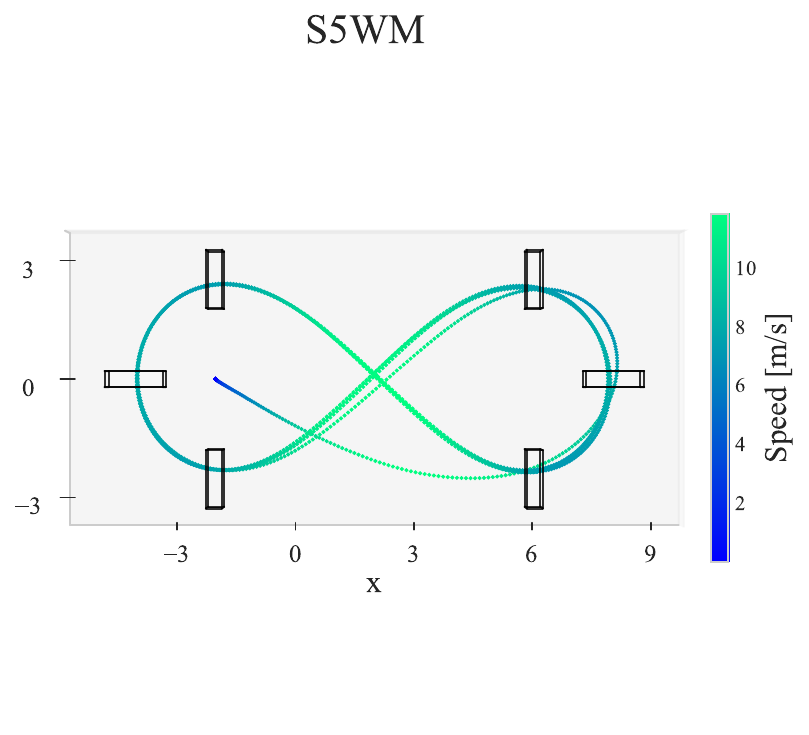}
    \end{minipage}%
    \begin{minipage}{0.3\textwidth}
        \centering
        \includegraphics[width=\linewidth, trim=20 10 60 0, clip]{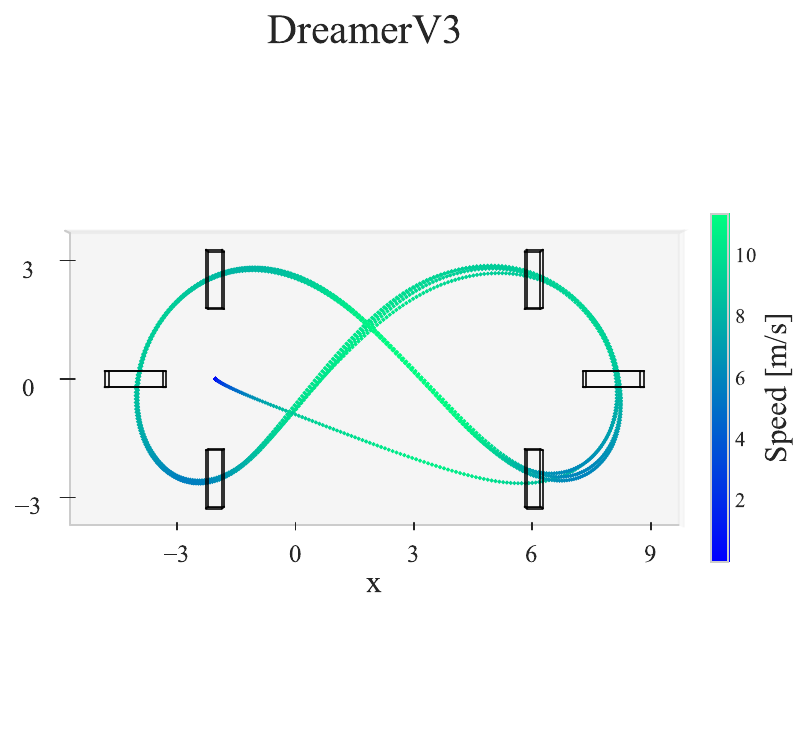}
    \end{minipage}%
    \begin{minipage}{0.35\textwidth}
        \centering
        \includegraphics[width=\linewidth, trim=20 10 0 0, clip]{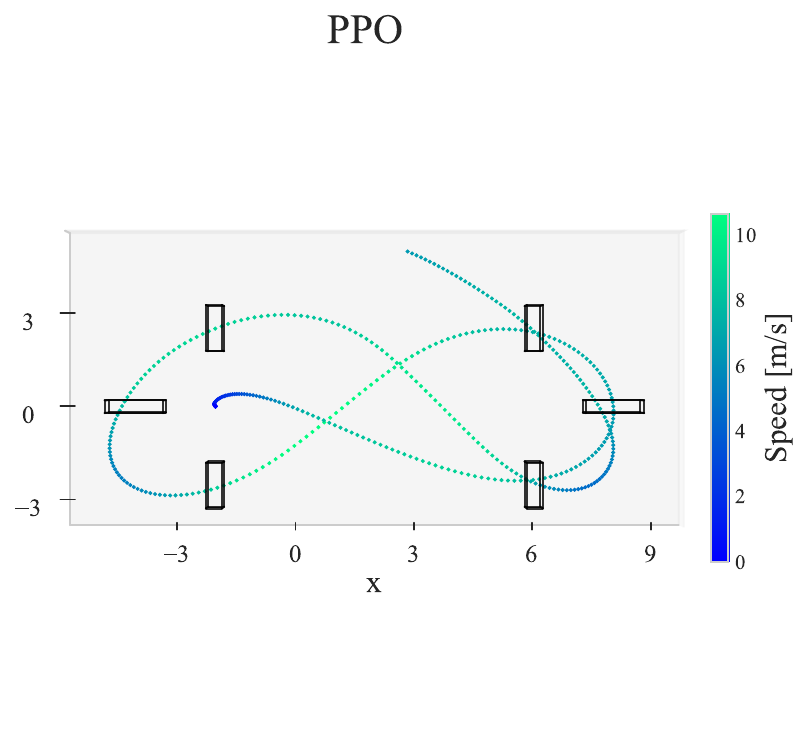}
    \end{minipage}%
\end{figure}
\vspace{-3em}
\begin{figure}[h]
    \label{fig:real_fig8}
    \begin{minipage}{0.95\textwidth}
    \caption{\textit{Vision-based Figure-8 task:} Comparison for S5WM (privileged), with baselines: DreamerV3 (privileged) and PPO (asymmetric). The asymmetric PPO crashes after the first lap and does not learn to consistently fly through the track.}
    \end{minipage}
\end{figure}
\newpage

\end{document}